%% file: main.tex
\documentclass[english,journal]{IEEEtran}
\usepackage{cancel}
\usepackage{amssymb}  %
\usepackage{mathtools}
\usepackage{import}
\usepackage{pgfplots}
\usepackage{color}
\usepackage{tabularx} %
\usepackage{multirow}
\usepackage{pgf}
\usepackage{blindtext}
\usepackage{cite}
\usepackage{siunitx}
\usepackage{balance}

\newcommand{\vect}[1]{\boldsymbol{#1}}

\renewcommand{\baselinestretch}{0.925}
\newlength{\bibitemsep}
\newlength{\bibparskip}
\let\oldthebibliography\thebibliography
\renewcommand\thebibliography[1]{%
  \oldthebibliography{#1}%
  \setlength{\parskip}{\bibitemsep}%
  \setlength{\itemsep}{\bibparskip}%
}

\makeatletter
\def\ps@headings{\def\@oddhead{\IEEEdoarxivheader{-1\oddsidemargin}\relax
\hbox{}\@IEEEheaderstyle\rightmark\hfil\thepage}\relax
\def\@evenhead{\IEEEdoarxivheader{-1\evensidemargin}\relax
\hbox{}\@IEEEheaderstyle\rightmark\hfil\thepage}\relax
\def\@oddfoot{\IEEEdoarxivfooter{-1\oddsidemargin}\hfil\hbox{}}\relax
\def\@evenfoot{\IEEEdoarxivfooter{-1\evensidemargin}\hfil\hbox{}}\relax}
\def\ps@IEEEtitlepagestyle{\ps@headings}

\def\IEEEarxivheadfootoffset{3pt}
\newdimen\IEEEheadtotopofpage
\newdimen\IEEEfoottobottomofpage
\newbox\@IEEEboxX

\def\IEEEarxivheader{}
\def\IEEEarxivfooter{}

\advance\IEEEheadtotopofpage by 1\topmargin
\advance\IEEEheadtotopofpage by 1in\relax

\IEEEfoottobottomofpage\paperheight
\advance\IEEEfoottobottomofpage by -1in\relax
\advance\IEEEfoottobottomofpage by -1\topmargin
\advance\IEEEfoottobottomofpage by -1\headheight
\advance\IEEEfoottobottomofpage by -1\headsep
\advance\IEEEfoottobottomofpage by -1\textheight
\advance\IEEEfoottobottomofpage by -1\footskip
\def\IEEEarxivheaderstyle{\normalfont\footnotesize}

\def\IEEEdoarxivheader#1{\@IEEEtrantmpdimenA\IEEEarxivheadfootoffset\relax
\@IEEEtrantmpdimenA -1\@IEEEtrantmpdimenA
\advance\@IEEEtrantmpdimenA by \IEEEheadtotopofpage
\settoheight{\@IEEEtrantmpdimenB}{\IEEEarxivheaderstyle HT}\relax
\advance\@IEEEtrantmpdimenA by -1\@IEEEtrantmpdimenB
\setbox\@IEEEboxX=\hbox{\relax
\raisebox{\@IEEEtrantmpdimenA}[0pt][0pt]{\parbox[t]{\textwidth}{\centering
\IEEEarxivheaderstyle\IEEEarxivheader}}}\relax
\wd\@IEEEboxX=0pt\relax
\ht\@IEEEboxX=0pt\relax
\dp\@IEEEboxX=0pt\relax
\box\@IEEEboxX\relax}

\def\IEEEarxivfooterstyle{\normalfont\footnotesize}

\def\IEEEdoarxivfooter#1{\@IEEEtrantmpdimenA\IEEEfoottobottomofpage\relax
\advance\@IEEEtrantmpdimenA by \IEEEarxivheadfootoffset\relax
\settodepth{\@IEEEtrantmpdimenB}{\IEEEarxivheaderstyle gjpqy}\relax
\advance\@IEEEtrantmpdimenA by 1\@IEEEtrantmpdimenB
\setbox\@IEEEboxX=\hbox{\hskip#1\hskip -1in\relax
\raisebox{\@IEEEtrantmpdimenA}[0pt][0pt]{\parbox[b]{\paperwidth}{\centering
\IEEEarxivfooterstyle\IEEEarxivfooter}}}\relax
\wd\@IEEEboxX=0pt\relax
\ht\@IEEEboxX=0pt\relax
\dp\@IEEEboxX=0pt\relax
\box\@IEEEboxX\relax}

\def\@IEEEheaderstyle{\normalfont\scriptsize}
\def\@IEEEfooterstyle{\normalfont\scriptsize}

\makeatother

\renewcommand{\IEEEarxivheadfootoffset}{3pt}
\renewcommand{\IEEEarxivheaderstyle}{\normalfont\footnotesize}
\renewcommand{\IEEEarxivfooterstyle}{\normalfont\scriptsize}
\renewcommand{\IEEEarxivheader}{This is the accepted version of an article
that has been published in 2022 IEEE/RSJ International Conference on Intelligent Robots and Systems (IROS). Changes were made to this version
by the publisher prior to publication.\\
The final version of record is available at 
\url{https://doi.org/10.1109/IROS47612.2022.9981928}}
\renewcommand{\IEEEarxivfooter}{ %
© 2022 IEEE.  Personal use of this material is permitted.  Permission from IEEE must be obtained for all other uses, in any current or future media, including reprinting/republishing this   material \\ for advertising or promotional purposes, creating new collective works, for resale or redistribution to servers or lists, or reuse of any copyrighted component of this work in other works}

\makeatother          
\begin{document}
\title{\LARGE \bf
BSA - Bi-Stiffness Actuation for optimally exploiting intrinsic compliance and inertial coupling effects in elastic joint robots
}
\author{Dennis Ossadnik, Mehmet C. Yildirim, Fan Wu, Abdalla Swikir, Hugo T. M. Kussaba, \\Saeed Abdolshah  and Sami Haddadin$^{*}$%
\thanks{$^{*}$Dennis Ossadnik, Mehmet C. Yildirim, Abdalla Swikir, Fan Wu, Hugo T. M. Kussaba, Saeed Abdolshah, and Sami Haddadin are with the Munich Institute of Robotics and Machine Intelligence (MIRMI), Technical University of Munich, Germany. Abdalla Swikir is also with the Department of Electrical and Electronic Engineering, Omar Al-Mukhtar University (OMU), Albaida, Libya
{\tt\small \{dennis.ossadnik, mehmet.yildirim, abdalla.swikir, f.wu, hugo.kussaba, saeed.abdolshah,  haddadin\}@tum.de}}%
}
\maketitle

\input{introduction}

\input{methods}
\input{friction_modeling}

\input{results}

\input{conclusion}

\vspace{-0.2cm}
\section*{Acknowledgement}
We gratefully acknowledge the funding support of Microsoft Germany, the Alfried Krupp von Bohlen and Halbach Foundation, and the European Union’s Horizon 2020 research and innovation programme as part of the project Darko under grant no. 101017274 and also under the Marie Skłodowska-Curie grant agreement no. 899987.

\bibliographystyle{ieeetr}

\end{document}

%% file: introduction.tex
\begin{abstract}
Compliance in actuation has been exploited to generate highly dynamic maneuvers such as throwing that take advantage of the potential energy stored in joint springs. However, the energy storage and release could not be well-timed yet. On the contrary, for multi-link systems, the natural system dynamics might even work against the actual goal. With the introduction of variable stiffness actuators, this problem has been partially addressed. With a suitable optimal control strategy, the approximate decoupling of the motor from the link can be achieved to maximize the energy transfer into the distal link prior to launch. However, such continuous stiffness variation is complex and typically leads to oscillatory swing-up motions instead of clear launch sequences. To circumvent this issue, we investigate decoupling for speed maximization with a dedicated  novel actuator concept denoted Bi-Stiffness Actuation. With this, it is possible to fully decouple the link from the joint mechanism by a switch-and-hold clutch and simultaneously keep the elastic energy stored. We show that with this novel paradigm, it is not only possible to reach the same optimal performance as with power-equivalent variable stiffness actuation, but even directly control the energy transfer timing. This is a major step forward compared to previous optimal control approaches, which rely on optimizing the full time-series control input.

\end{abstract}
\section{Introduction}
The capability of biological muscles to serve as both motors and springs is the reason for their high energy efficiency and performance compared to the commonly used rigid actuators in robotics. There have been efforts to reach similar performance in the robotics community by introducing elastic elements in the drive-train \cite{Zinn04}. For example, series elastic actuators (SEAs) \cite{Pratt1995} implement {an elastic component with} fixed stiffness in series to the motor. 

Due to their capability to store and release energy in the spring, SEAs are capable of highly dynamic maneuvers, outperforming the rigid actuator. 
Previous works investigating explosive movements \cite{Haddadin2011,Haddadin2012} show that
the control signals for maximizing the end-link velocity are of bang-bang type \cite{haddadin2009kick}, resulting in a resonant excitation. However, this strategy cannot be observed in biological systems. Players of ball sports, for example, utilize their muscular strength in a very fast, coordinated stroking movement \cite{grezios2006muscle}.
\begin{figure}[h]%
	\centering
	\def\svgwidth{0.4\textwidth}
	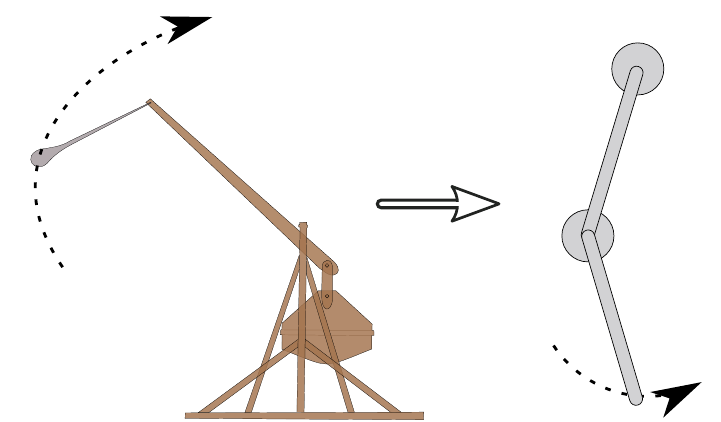
	\caption[Model overview]{By taking advantage of inertia timing and decoupling, we can precisely control the timing of energy transfer in an elastic double pendulum similar to launching a trebuchet.}
	\label{fig:trebuchet}
	\vspace*{-0.65cm}
\end{figure}

In handball, the throwing movement involves a sequential action of proximal to distal body segments \cite{joris1985force}. In trained individuals, the muscles are only active for a small fraction of the total motion duration. Each segment in the kinematic chain has to contribute at exactly the right time to transfer the energy to the distal parts of the chain. This \textit{inertia timing} might be the key to understanding the optimal coordination of fast movements. As Lehnertz points out in~\cite{lehnertz1984molekularmechanische}, ``the available muscle strength can only be converted into optimal movement speed by making the best possible use of the inertial forces that occur.'' In SEAs, such a sequence is impossible, since ``the timing of energy storage and release
is not independent'' \cite{Plooij}. 

With the introduction of variable stiffness actuators (VSAs), which employ a second actuator for stiffness adaptation \cite{vanderborght2013variable, MACCEPA, DLR_VS,Eiberger2010}, this problem could be partially addressed. 
The inertia timing phenomenon was investigated in \cite{Haddadin2012}, where the end-link velocity of a double pendulum driven by VSAs was maximized by leveraging energy storage and release. The problem was formulated as an optimization problem, and its solution resulted in bang-bang-like signals of both the motor velocity and spring stiffness. 
It is important to mention that those signals were generated due to the ability to set the stiffness to zero, which in turn made it possible to fully decouple the link side. The result holds for low final times in the optimization problem. Using a higher final time, again resonant excitation can be dominant, similar to the SEA case. 

Recently, another highly promising type of actuation has emerged: Clutched elastic actuators (CEAs) \cite{Plooij} employ clutches in the drivetrain to for example lock or bypass springs \cite{Chen2013, leach2013linear}. With a suitable choice of clutch mechanism, it is possible to directly control the timing of energy storage and release.  
\vspace{-0.1cm}
\subsection{Contribution}
To address the issue of VSAs being unable to directly control the energy transfer timing, we propose a novel \textit{Bi-Stiffness Actuation} (BSA) concept based on a clutch mechanism and investigate how inertial decoupling can be optimally exploited for explosive movements. By encoding a proximal-distal mode sequence inspired by biological systems, the optimal control problem for such a hybrid system is simplified and allows direct control of the transition timing. A double pendulum equipped with our actuator thus can be launched like a trebuchet (Fig. \ref{fig:trebuchet}). Numerical experiments were performed to (1) maximize end-link velocity and (2) minimize control efforts for a fixed target velocity. The results demonstrate that  
\begin{itemize}
    \item Our actuator is capable of precisely controlling the timing of energy storage and release,
    \item By ensuring both VSA and BSA have the same power input, we reach similar performance in terms of the maximum end-link velocity,
    \item We avoid resonant excitation and always obtain an optimal control solution that results in a catapult-like rapid conversion of potential energy.
\end{itemize}
The paper is organized as follows: In Section~\ref{scc:model}, we introduce the dynamic model of the proposed actuator concept and restate the dynamics of the previously suggested VSA model. Then, in Section~\ref{scc:evaluation}, we evaluate the actuator and give a conclusion in Section~\ref{scc:conc}.

%% file: figures/trebuchet.pdf_tex
\begingroup%
  \makeatletter%
  \providecommand\color[2][]{%
    \errmessage{(Inkscape) Color is used for the text in Inkscape, but the package 'color.sty' is not loaded}%
    \renewcommand\color[2][]{}%
  }%
  \providecommand\transparent[1]{%
    \errmessage{(Inkscape) Transparency is used (non-zero) for the text in Inkscape, but the package 'transparent.sty' is not loaded}%
    \renewcommand\transparent[1]{}%
  }%
  \providecommand\rotatebox[2]{#2}%
  \newcommand*\fsize{\dimexpr\f@size pt\relax}%
  \newcommand*\lineheight[1]{\fontsize{\fsize}{#1\fsize}\selectfont}%
  \ifx\svgwidth\undefined%
    \setlength{\unitlength}{207.50546193bp}%
    \ifx\svgscale\undefined%
      \relax%
    \else%
      \setlength{\unitlength}{\unitlength * \real{\svgscale}}%
    \fi%
  \else%
    \setlength{\unitlength}{\svgwidth}%
  \fi%
  \global\let\svgwidth\undefined%
  \global\let\svgscale\undefined%
  \makeatother%
  \begin{picture}(1,0.58498896)%
    \lineheight{1}%
    \setlength\tabcolsep{0pt}%
    \put(0,0){\includegraphics[width=\unitlength,page=1]{./figures/trebuchet.pdf}}%
  \end{picture}%
\endgroup%

%% file: methods.tex
\section{Modeling}
\label{scc:model}
In this section, we briefly outline the concept for a single BSA actuator. Then, we derive a hybrid system model for an elastic double pendulum that is actuated by either two BSAs or two VSAs and discuss the power flow for each system. Additionally, we derive a frictional contact model for the BSA to account for transient slipping dynamics that might occur in a physical prototype of the system. 

\subsection{BSA Concept}
\begin{figure}[h]
	\centering
	\def\svgwidth{0.5\textwidth}
	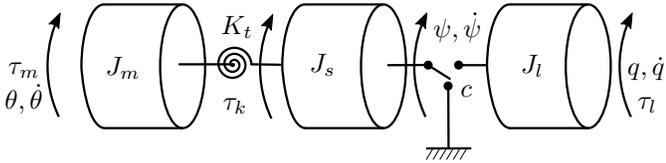 \vspace*{-1cm}
	\caption[Model overview]{Sketch of the proposed actuator. The spring inertia can be locked in place, while the link is decoupled or connected directly to the link. We can therefore distinguish two distinct modes: The decoupled mode (DEC) and the series elastic actuator mode (SEA). }
	\label{fig:Model}
	
\end{figure}

Our concept consists of a motor that is connected to a spring, similar to a series elastic actuator (SEA). The spring inertia is connected to a switch-and-hold mechanism $c$, which either brakes the spring or connects it to the link-side. A sketch of the system can be seen in Fig. \ref{fig:Model}. The explicit modeling of the spring inertia is an important aspect of this model, which will become evident in our hybrid system formulation later in this section.
\vspace{-0.65cm}

To simplify the formulation of the optimal control problem, we model the motor as an ideal velocity source. Instead of the motor torque $\tau_m$, we now assume the motor velocity $\dot{\theta}$ as an input. The spring inertia is subject to the spring torque $\tau_k = K_t (\theta - \psi)$, where $K_t$, $\theta$, and $\psi$ are the spring stiffness, the motor angle, and spring output angle, respectively. %

\subsection{Double pendulum actuated by two BSAs} \label{sec:BSA}
\begin{figure}
\centering
\vspace{0.25cm}
\def\svgwidth{0.25\textwidth}
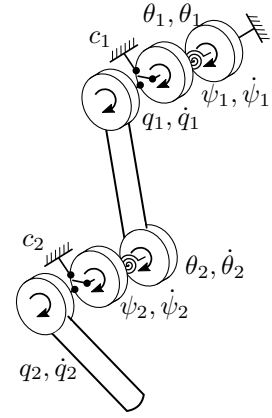 
\caption{Sketch of the 2-DoF pendulum actuated by two BSAs. The double pendulum has four modes in total, which depend on the state of the switching clutches $c_1$ and $c_2$. Each joint is either in SEA or DEC mode.} 
\label{fig:arm}
\vspace{-0.45cm}
\end{figure}

The decoupling capability of our actuator becomes especially important when dealing with nonlinear dynamics, where inertial coupling is present. As a simple example for such a system, we are going to analyse a double pendulum  driven by two BSAs, see Fig.~\ref{fig:arm}.

We employ Spongs' assumption that the coupling between actuator and link acts solely through the link torque $\tau_l$ \cite{spong1989modeling}. By introducing the vector of spring $\vect{\psi} = [\psi_1, \psi_2]^{\mathsf{T}}$ and link positions $\vect{q} = [q_1, q_2]^{\mathsf{T}}$ and defining $\vect{\xi} \coloneqq [\vect{\psi}, \vect{q}]^{\mathsf{T}}\in \mathbb{R}^{4}$, the dynamics of the system can be formulated  as
\begin{gather}
    \underbrace{
    \begin{bmatrix} 
	        \vect{B} & \vect{0} \\
	        \vect{0}  & \vect{M}(\vect{q})
	\end{bmatrix}}_{\eqqcolon \vect{\Pi}(\vect{\xi})} \ddot{\vect{\xi}} + \underbrace{
    \begin{bmatrix}
        \vect{0} \\
	    \vect{h}(\vect{q}, \dot{\vect{q}})
	\end{bmatrix}}_{\eqqcolon \vect{\eta}(\vect{\xi}, \dot{\vect{\xi}})} + \underbrace{
	\begin{bmatrix}
        \vect{K} (\vect{\theta} - \vect{\psi}) \\
        \vect{0}
    \end{bmatrix}}_{\eqqcolon \vect{\tau}_k} = \vect{C}_p^{\mathsf{T}} \vect{\lambda}, 
    \label{eq:robot}\\
    \vect{C}_p \dot{\vect{\xi}} = \vect{0}, 
    \label{eq:constraint}
\end{gather}
where $\vect{B} = \mathsf{diag}([J_s, J_s])$, $\vect{M}(\vect{q})$, $\vect{K}\in \mathbb{R}^{2\times2}$, $\vect{h}(\vect{q}, \dot{\vect{q}})\in \mathbb{R}^{2}$ is the spring inertia matrix, the link-side inertia matrix, the stiffness matrix, and the link-side nonlinear bias term, respectively.  The vector $\vect{\lambda}\in \mathbb{R}^{4}$ is the constraint torque, and $\vect{\theta}= [\theta_1, \theta_2]^{\mathsf{T}}$ is the motor position obtained by integrating the motor velocity, denoted by $\dot{\vect{\theta}}=[\dot{\theta}_1, \dot{\theta}_2]^{\mathsf{T}}$. Similar to the single actuator, we assume that the motor is an ideal velocity source. The matrix $\vect{C}_p$ defines the switching through the switch-and-hold mechanism. More precisely, we use the clutch to prevent the relative motion between the spring and the link-side \cite{Plooij}. If the switch-and-hold mechanism brakes the spring inertia, $\dot{\psi}  = 0$ must hold. If it connects the spring inertia to the link-side, the spring and link velocities must match, i.e. $\dot{\psi}  = \dot{q} $. These conditions can be easily encoded into the oriented incidence matrix $\vect{C}_p$. Since there are two actuators with one switch-and-hold mechanism each, there are in total four modes for the double pendulum. The constraint matrices for each mode can be found in Table \ref{tab:modes}. 
\begin{table}
\vspace{0.25cm}
\caption{Actuator modes. If $c_i = 0$ the switch is connected to the ground, else if  $c_i = 1$ it is connected to the link.}
\centering
	\begin{tabular}[c]{|c|c|c|c|c|}
	\hline	\textbf{Mode} & $p$& $c_1$ & $c_2$ & \textbf{Constraint Matrix} $\vect{C}$\\
	\hline \rule{0pt}{4ex}    DEC-DEC &1 &0& 0& $ \begin{bmatrix}
	1 & 0& 0& 0 \\
	0& 1& 0& 0
	\end{bmatrix}$        ~\\[0.25cm]
	\hline	\rule{0pt}{4ex}    SEA-SEA &2&1& 1& $\begin{bmatrix}
	1& 0& -1& 0 \\
	0& 1& 0& -1
	\end{bmatrix}$ ~\\[0.25cm]
	\hline\rule{0pt}{4ex}    	DEC-SEA&3&0& 1&  $\begin{bmatrix}
	1 & 0& 0& 0 \\
	0& 1& 0& -1 
	\end{bmatrix}$  ~\\[0.25cm]
	\hline\rule{0pt}{4ex}    	SEA-DEC&4&1& 0&  $ \begin{bmatrix}
	1& 0& -1& 0\\
	0 & 1& 0& 0 
	\end{bmatrix}$ ~\\[0.25cm] \hline
\end{tabular}
\vspace{-0.25cm} \label{tab:modes}
\end{table}

In order to continue modeling the proposed mechanism as a hybrid system, there are two main points that need to be taken into account, as follows.

\textbf{Constraint torque elimination.} The following derivation closely follows \cite{Sobotka2007}. For brevity, the dependency on $\vect{\xi}$ and $\dot{\vect{\xi}}$ is omitted. Differentiating \eqref{eq:constraint} yields
\begin{equation}
\vect{C}_p \ddot{\vect{\xi}} + \cancel{\dot{\vect{C}_p} \dot{\vect{\xi}}} = \vect{0}. \notag
\end{equation}
We now left-multiply \eqref{eq:robot} by $ \vect{C}_p \vect{\vect{\Pi}}^{-1}$, which leads to
\begin{equation}
\cancel{\vect{C}_p  \ddot{\vect{\xi}}} + \vect{C}_p \vect{\Pi}^{-1} \vect{\tau}_k + \vect{C}_p \vect{\Pi}^{-1} \vect{\eta} =   \vect{C}_p \vect{\Pi}^{-1} \vect{C}_p^{\mathsf{T}} \vect{\lambda}. \notag
\end{equation}
We can now directly calculate the constraint torques and eliminate them from the equations of motion:
\begin{equation} \label{eq:constr_torque}
		 \vect{\lambda} = (\vect{C}_p \vect{\Pi}^{-1} \vect{C}_p^{\mathsf{T}})^{-1}\vect{C}_p \vect{\Pi}^{-1} ( \vect{\tau}_k + \vect{\eta}).
\end{equation} 
Here, we denoted the constraint \textcolor{black}{torques'} dependence $\vect{C}_p$ by adding a subscript $p$.

\textbf{Impact analysis.} The system exhibits an instantaneous jump in the velocities upon mode transition. \textcolor{black}{Similar to \cite{Sobotka2007, ossadnik2021nonlinear, ossadnik2021ult, westervelt2018feedback}, we assume that
\begin{itemize}
    \item[A1)] an impact occurs whenever there is a change in contact situation (i.e. toggling the switch-and-hold mechanism);
\item[A2)]  the impact is instantaneous;
\item[A3)]  the forces due to the impact can be represented by impulses;
\item[A4)]  the motors do not generate impulses and do not have to be considered in the analysis; and,
\item[A5)]  there is no discontinuity in the positions, but an instantaneous change in the actuators’ velocities.
\end{itemize} 
Under these assumptions, the following equation between differential measures can be obtained \cite{leine2007stability}
\begin{equation}
\boldsymbol{\Pi}(\boldsymbol{\xi})\,\mathrm{d}\dot{\boldsymbol{\xi}}+\boldsymbol{\eta}(\boldsymbol{\xi}, \dot{\boldsymbol{\xi}}) \,\mathrm{d} t= -\boldsymbol{\tau}_k\,\mathrm{d}t+\boldsymbol{C}_p^{\mathsf{T}} \mathrm{d}\boldsymbol{\lambda}_p,  \label{eq:coll_full} 
\end{equation}
with the differential measures $\mathrm{d}\dot{\boldsymbol{\xi}}$ and $\mathrm{d}\boldsymbol{\lambda}_p$ satisfying 
\begin{align} %
    \mathrm{d}\dot{\vect{\xi}} &= \ddot{\vect{\xi}}\,\mathrm{d}t + (\dot{\vect{\xi}}^{+} - \dot{\vect{\xi}}^{-})\,\mathrm{d}\nu, \notag \\
    \mathrm{d}{\vect{\lambda}}_p &= {\vect{\lambda}}_p\,\mathrm{d}t + \vect{\Lambda}_p\,\mathrm{d} \nu, \notag
\end{align}
where $\dot{\vect{\xi}}^{+} \coloneqq \lim_{t\rightarrow t_c^{+}}\dot{\boldsymbol{\xi}}(t)$, $\dot{\vect{\xi}}^{-} \coloneqq \lim_{t\rightarrow t_c^{-}}\dot{\boldsymbol{\xi}}(t)$, $\mathrm{d}\nu$ is the atomic measure, and $\vect{\Lambda}_p$ is the contact impulse \cite{leine2007stability}. We refer interested readers to \cite{brogliato1996nonsmooth,leine2007stability} for detailed information on differential measures and the atomic measure used in this paper.
} %

\textcolor{black}{
Integrating \eqref{eq:coll_full} over an instant $t_c$ where the impulse $\vect{\Lambda}_p$ is non-zero 
yields the conservation of momentum equation \cite{leine2007stability}
$$
\boldsymbol{\Pi}(\boldsymbol{\xi})
(\dot{\vect{\xi}}^{+} - \dot{\vect{\xi}}^{-})=\boldsymbol{C}_p^{\mathsf{T}} \vect{\Lambda}_p. \label{eq:coll_eval}
$$
} %
Since $\vect{C}_p\dot{\vect{\xi}}^{+} = \vect{0}$ must hold, we can solve for the impulse directly and obtain an expression for the updated velocity
\begin{align}\label{eq:impact_law}
	\dot{\vect{\xi}}^{+} &= \dot{\vect{\xi}}^{-} +  \vect{\Pi}^{-1} \vect{C}_p^{\mathsf{T}} \vect{\Lambda}_p, \\ 
	\vect{\Lambda}_p &= -(\vect{C}_p \vect{\Pi}^{-1} \vect{C}_p^{\mathsf{T}})^{-1} \vect{C}_p \dot{\vect{\xi}}^{-} \notag.   
\end{align}
Now, we have all the ingredients to formally describe the double pendulum system actuated by two BSAs as a hybrid system.

\textbf{Hybrid system model. } The overall dynamics of our system is characterized by the interaction of continuous and discrete dynamics. The switching between the modes is controlled by the user, and the velocity is reset after switching according to equation \eqref{eq:impact_law}. Formally, such a system can be described as a switched system with impulsive effects \cite{liberzon2003switching}, which is a special form of a hybrid system. We define the continuous state $\vect{x} \coloneqq [\vect{\theta}, \vect{\xi}, \dot{\vect{\xi}}]^{\mathsf{T}} \in \mathbb{R}^{10}$, the control input\footnote{The index $\theta$ is used to emphasize that the control input is the motor velocity and to distinguish it from the stiffness adjustment control that comes later in Section \ref{sec:vsa}.} $\vect{u} = \vect{u}_{\theta} \in \mathbb{R}^2$ and the index set $\mathcal{P} = \{1, 2, 3, 4\}$. Now, the state space dynamics of the system for the modes $p \in \mathcal{P}$ can be formulated as
\begin{equation}
	\dot{\vect{x}} = \vect{f}_p(\vect{x}, \vect{u}) \coloneqq \begin{bmatrix}
	\vect{u}_{\theta} \\
	\dot{\vect{\xi}} \\
	 \vect{\Pi}^{-1} (\vect{C}_p^{\mathsf{T}} \vect{\lambda}_p - \vect{\eta} - \vect{\tau}_k)
	\end{bmatrix}. \label{eq:dyn}
\end{equation}
We impose the usual regularity assumptions on, $\vect{f}_p$ such that the solutions of the differential equation  \eqref{eq:dyn}
are well-defined \cite{Angeli}. Moreover, we define the jump map that describes the discrete dynamics governing the transition to a mode $p \in \mathcal{P}$ as
\begin{equation}
\vect{x}^{+} = \vect{g}_p(\vect{x}^{-}) \coloneqq \begin{bmatrix}
\vect{\theta}^{-} \\
\vect{\xi}^{-} \\
\dot{\vect{\xi}}^{-} + \vect{\Pi}^{-1} \vect{C}_p^{\mathsf{T}} \vect{\Lambda}_p \\
\end{bmatrix}. \label{eq:jump}
\end{equation}
We also define a switching signal $\sigma: \mathbb{R}^{+} \rightarrow \mathcal{P}$ such that, given the family in equation \eqref{eq:dyn} and \eqref{eq:jump}, a switched-impulsive system
\begin{align}
	\dot{\vect{x}} = \vect{f}_{\sigma}(\vect{x}, \vect{u}),~
	\vect{x}^{+} = \vect{g}_{\sigma}(\vect{x}^{-})\label{eq:dyn1}
\end{align}
is generated. Please note that, in our setting, the switching signal $\sigma$ is another control signal. In other words, the user determines when switching should take place, i.e. when the clutches should be opened and closed.

For sake of completeness and later comparison, we will briefly go over the process of modeling the VSA system in the following subsection.   
\subsection{VSA Modeling} \label{sec:vsa}

We proceed to model the VSA system. The equations of motion for this system are given by
\begin{equation}
	\vect{M}(\vect{q}) \ddot{\vect{q}} +\vect{h}(\vect{q}, \dot{\vect{q}})   + \vect{K}(\vect{\theta} - \vect{q})  = \vect{0}.
\end{equation}
Again, we model the motor as an ideal velocity source. Here, $\vect{K} = \mathsf{diag}(\vect{k})$, with the individual stiffness values being summarized in a vector $\vect{k} = [K_1, K_2]^{\mathsf{T}}$. The stiffness values can be dynamically changed by a stiffness adjusting mechanism. Instead of explicitly modeling this mechanism, we consider the velocity of the stiffness adjustment as another control input. Defining the state \mbox{$\vect{x} \coloneqq [\vect{\theta}, \vect{k}, \vect{q}, \dot{\vect{q}}]^{\mathsf{T}} \in \mathbb{R}^8$} and the control input $\vect{u} = [\vect{u}_{\theta}, \vect{u}_k]^{\mathsf{T}} \in \mathbb{R}^4$, the state space dynamics of the VSA can now be formulated as
\begin{equation}
	\dot{\vect{x}} = \vect{f}(\vect{x}, \vect{u}) \coloneqq \begin{bmatrix}
	\vect{u}_{\theta} \\
		\vect{u}_{k} \\
		\dot{\vect{q}} \\
	\vect{M}^{-1}(	-\vect{h}  -\vect{K}(\vect{\theta} - \vect{q}))
	\end{bmatrix}.
\end{equation}
In addition to the system dynamics, we elaborate in the next section on the differences in the power flow between BSA and VSA.  
\subsection{Power flow}
\begin{figure}[h]
	\centering
	\vspace*{-1cm}
	\def\svgwidth{0.45\textwidth}
	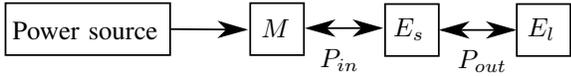
	\vspace*{-1cm}
	\caption[Model overview]{Power flow of the BSA actuator. The only source of power is the motor $M$ for changing the equilibrium position of the spring. }\label{fig:powerDCEA}
\end{figure}

\begin{figure}[h]
	\centering
	\def\svgwidth{0.45\textwidth}
	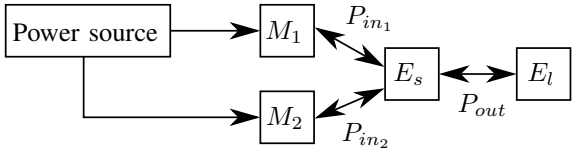
	\vspace*{-0.6cm}
	\caption[Model overview]{Power flow of the VSA actuator. There are two sources of power: The motor $M_1$ for changing the equilibrium position of the spring and the motor $M_2$ for the stiffness adjustment mechanism.}\label{fig:powerVSA}
\end{figure}

For comparing the energy efficiency of both actuators, we have to consider the power flow from the motors into the spring as well as the power flow from the spring into the link-side. A schematic overview of the power flow for both systems can be seen in Fig. \ref{fig:powerDCEA} and \ref{fig:powerVSA} (drawn similar to \cite{wu2020}). Since no damping is assumed, the power supplied through each spring $i$ (for brevity we omit the index in the following) can be computed as 
\begin{equation}
    P_{out} = \tau_{k} \dot{q}. \notag
\end{equation}
In the BSA case, the spring torque is given as $\tau_{k} = K (\theta - \psi)$, and in the \textcolor{black}{VSA} case, it is $\tau_{k} = K (\theta - q)$. As Braun et al. describe in \cite{braun2013robots}, the total power flow into the spring can be computed as
\begin{equation}
P_{in} = P_{out} + \dot{E}_s, \label{eq:power} \notag
\end{equation}
where $\dot{E}_s$ is the time derivative of the spring energy. In case of the BSA, this term can be directly computed as
\begin{equation}
    \dot{E}_s = K (\theta - \psi) (\dot{\theta} - \dot{\psi}), \notag
\end{equation}
while in the VSA case, we have to explicitly consider the time derivative of the stiffness adjustment, i.e.
\begin{equation}
    \dot{E}_s = \frac{1}{2}  \dot{K} (\theta - q)^2 + K (\theta - q) (\dot{\theta} - \dot{q}). \notag
\end{equation}

\vspace{-0.65cm}

%% file: figures/clutch_system_switch.pdf_tex
\begingroup%
  \makeatletter%
  \providecommand\color[2][]{%
    \errmessage{(Inkscape) Color is used for the text in Inkscape, but the package 'color.sty' is not loaded}%
    \renewcommand\color[2][]{}%
  }%
  \providecommand\transparent[1]{%
    \errmessage{(Inkscape) Transparency is used (non-zero) for the text in Inkscape, but the package 'transparent.sty' is not loaded}%
    \renewcommand\transparent[1]{}%
  }%
  \providecommand\rotatebox[2]{#2}%
  \newcommand*\fsize{\dimexpr\f@size pt\relax}%
  \newcommand*\lineheight[1]{\fontsize{\fsize}{#1\fsize}\selectfont}%
  \ifx\svgwidth\undefined%
    \setlength{\unitlength}{227.41072046bp}%
    \ifx\svgscale\undefined%
      \relax%
    \else%
      \setlength{\unitlength}{\unitlength * \real{\svgscale}}%
    \fi%
  \else%
    \setlength{\unitlength}{\svgwidth}%
  \fi%
  \global\let\svgwidth\undefined%
  \global\let\svgscale\undefined%
  \makeatother%
  \begin{picture}(1,0.39104827)%
    \lineheight{1}%
    \setlength\tabcolsep{0pt}%
    \put(0,0){\includegraphics[width=\unitlength,page=1]{./figures/clutch_system_switch.pdf}}%
    \put(0.017906,0.19953136){\color[rgb]{0,0,0}\makebox(0,0)[lt]{\lineheight{1.25}\smash{\begin{tabular}[t]{l}$\tau_m$\end{tabular}}}}%
    \put(0.01319454,0.14841241){\color[rgb]{0,0,0}\makebox(0,0)[lt]{\lineheight{1.25}\smash{\begin{tabular}[t]{l}$\theta, \dot{\theta}$\end{tabular}}}}%
    \put(0.1606622,0.19776457){\color[rgb]{0,0,0}\makebox(0,0)[lt]{\lineheight{1.25}\smash{\begin{tabular}[t]{l}$J_m$\end{tabular}}}}%
    \put(0.45995546,0.20188707){\color[rgb]{0,0,0}\makebox(0,0)[lt]{\lineheight{1.25}\smash{\begin{tabular}[t]{l}$J_s$\end{tabular}}}}%
    \put(0.33014612,0.25810366){\color[rgb]{0,0,0}\makebox(0,0)[lt]{\lineheight{1.25}\smash{\begin{tabular}[t]{l}$K_t$\end{tabular}}}}%
    \put(0,0){\includegraphics[width=\unitlength,page=2]{./figures/clutch_system_switch.pdf}}%
    \put(0.92591586,0.19762293){\color[rgb]{0,0,0}\makebox(0,0)[lt]{\lineheight{1.25}\smash{\begin{tabular}[t]{l}$q, \dot{q}$\end{tabular}}}}%
    \put(0,0){\includegraphics[width=\unitlength,page=3]{./figures/clutch_system_switch.pdf}}%
    \put(0.76983354,0.20038455){\color[rgb]{0,0,0}\makebox(0,0)[lt]{\lineheight{1.25}\smash{\begin{tabular}[t]{l}$J_l$\end{tabular}}}}%
    \put(0,0){\includegraphics[width=\unitlength,page=4]{./figures/clutch_system_switch.pdf}}%
    \put(0.64132102,0.25332066){\color[rgb]{0,0,0}\makebox(0,0)[lt]{\lineheight{1.25}\smash{\begin{tabular}[t]{l}$\psi, \dot{\psi}$\end{tabular}}}}%
    \put(0,0){\includegraphics[width=\unitlength,page=5]{./figures/clutch_system_switch.pdf}}%
    \put(0.94067044,0.14749121){\color[rgb]{0,0,0}\makebox(0,0)[lt]{\lineheight{1.25}\smash{\begin{tabular}[t]{l}$\tau_l$\end{tabular}}}}%
    \put(0,0){\includegraphics[width=\unitlength,page=6]{./figures/clutch_system_switch.pdf}}%
    \put(0.33283857,0.14647326){\color[rgb]{0,0,0}\makebox(0,0)[lt]{\lineheight{1.25}\smash{\begin{tabular}[t]{l}$\tau_k$\end{tabular}}}}%
    \put(0,0){\includegraphics[width=\unitlength,page=7]{./figures/clutch_system_switch.pdf}}%
    \put(0.68137224,0.16500222){\color[rgb]{0,0,0}\makebox(0,0)[lt]{\lineheight{1.25}\smash{\begin{tabular}[t]{l}$c$\end{tabular}}}}%
  \end{picture}%
\endgroup%

%% file: figures/arm.pdf_tex
\begingroup%
  \makeatletter%
  \providecommand\color[2][]{%
    \errmessage{(Inkscape) Color is used for the text in Inkscape, but the package 'color.sty' is not loaded}%
    \renewcommand\color[2][]{}%
  }%
  \providecommand\transparent[1]{%
    \errmessage{(Inkscape) Transparency is used (non-zero) for the text in Inkscape, but the package 'transparent.sty' is not loaded}%
    \renewcommand\transparent[1]{}%
  }%
  \providecommand\rotatebox[2]{#2}%
  \newcommand*\fsize{\dimexpr\f@size pt\relax}%
  \newcommand*\lineheight[1]{\fontsize{\fsize}{#1\fsize}\selectfont}%
  \ifx\svgwidth\undefined%
    \setlength{\unitlength}{91.60713977bp}%
    \ifx\svgscale\undefined%
      \relax%
    \else%
      \setlength{\unitlength}{\unitlength * \real{\svgscale}}%
    \fi%
  \else%
    \setlength{\unitlength}{\svgwidth}%
  \fi%
  \global\let\svgwidth\undefined%
  \global\let\svgscale\undefined%
  \makeatother%
  \begin{picture}(1,1.2514621)%
    \lineheight{1}%
    \setlength\tabcolsep{0pt}%
    \put(0,0){\includegraphics[width=\unitlength,page=1]{./figures/arm.pdf}}%
    \put(0.38844938,1.16775183){\color[rgb]{0,0,0}\makebox(0,0)[lt]{\lineheight{1.25}\smash{\begin{tabular}[t]{l}$\theta_1, \dot{\theta}_1$\end{tabular}}}}%
    \put(0.38556012,0.86444475){\color[rgb]{0,0,0}\makebox(0,0)[lt]{\lineheight{1.25}\smash{\begin{tabular}[t]{l}$q_1, \dot{q}_1$\end{tabular}}}}%
    \put(0.55969805,0.93206037){\color[rgb]{0,0,0}\makebox(0,0)[lt]{\lineheight{1.25}\smash{\begin{tabular}[t]{l}$\psi_1, \dot{\psi}_1$\end{tabular}}}}%
    \put(0.22987905,1.10728883){\color[rgb]{0,0,0}\makebox(0,0)[lt]{\lineheight{1.25}\smash{\begin{tabular}[t]{l}$c_1$\end{tabular}}}}%
    \put(0.03412738,0.50992913){\color[rgb]{0,0,0}\makebox(0,0)[lt]{\lineheight{1.25}\smash{\begin{tabular}[t]{l}$c_2$\end{tabular}}}}%
    \put(0.51261871,0.42785736){\color[rgb]{0,0,0}\makebox(0,0)[lt]{\lineheight{1.25}\smash{\begin{tabular}[t]{l}$\theta_2, \dot{\theta}_2$\end{tabular}}}}%
    \put(0.02569991,0.13034649){\color[rgb]{0,0,0}\makebox(0,0)[lt]{\lineheight{1.25}\smash{\begin{tabular}[t]{l}$q_2, \dot{q}_2$\end{tabular}}}}%
    \put(0.32030142,0.31002452){\color[rgb]{0,0,0}\makebox(0,0)[lt]{\lineheight{1.25}\smash{\begin{tabular}[t]{l}$\psi_2, \dot{\psi}_2$\end{tabular}}}}%
    \put(0,0){\includegraphics[width=\unitlength,page=2]{./figures/arm.pdf}}%
  \end{picture}%
\endgroup%

%% file: figures/powerDCEA.pdf_tex
\begingroup%
  \makeatletter%
  \providecommand\color[2][]{%
    \errmessage{(Inkscape) Color is used for the text in Inkscape, but the package 'color.sty' is not loaded}%
    \renewcommand\color[2][]{}%
  }%
  \providecommand\transparent[1]{%
    \errmessage{(Inkscape) Transparency is used (non-zero) for the text in Inkscape, but the package 'transparent.sty' is not loaded}%
    \renewcommand\transparent[1]{}%
  }%
  \providecommand\rotatebox[2]{#2}%
  \newcommand*\fsize{\dimexpr\f@size pt\relax}%
  \newcommand*\lineheight[1]{\fontsize{\fsize}{#1\fsize}\selectfont}%
  \ifx\svgwidth\undefined%
    \setlength{\unitlength}{184.50117216bp}%
    \ifx\svgscale\undefined%
      \relax%
    \else%
      \setlength{\unitlength}{\unitlength * \real{\svgscale}}%
    \fi%
  \else%
    \setlength{\unitlength}{\svgwidth}%
  \fi%
  \global\let\svgwidth\undefined%
  \global\let\svgscale\undefined%
  \makeatother%
  \begin{picture}(1,0.36992274)%
    \lineheight{1}%
    \setlength\tabcolsep{0pt}%
    \put(0,0){\includegraphics[width=\unitlength,page=1]{./figures/powerDCEA.pdf}}%
    \put(0.05389293,0.17949931){\color[rgb]{0,0,0}\makebox(0,0)[lt]{\lineheight{1.25}\smash{\begin{tabular}[t]{l}Power source\end{tabular}}}}%
    \put(0,0){\includegraphics[width=\unitlength,page=2]{./figures/powerDCEA.pdf}}%
    \put(0.45925098,0.18184988){\color[rgb]{0,0,0}\makebox(0,0)[lt]{\lineheight{1.25}\smash{\begin{tabular}[t]{l}$M$\end{tabular}}}}%
    \put(0.67450174,0.18306243){\color[rgb]{0,0,0}\makebox(0,0)[lt]{\lineheight{1.25}\smash{\begin{tabular}[t]{l}$E_s$\end{tabular}}}}%
    \put(0.89265902,0.18306237){\color[rgb]{0,0,0}\makebox(0,0)[lt]{\lineheight{1.25}\smash{\begin{tabular}[t]{l}$E_l$\end{tabular}}}}%
    \put(0.55495067,0.13291991){\color[rgb]{0,0,0}\makebox(0,0)[lt]{\lineheight{1.25}\smash{\begin{tabular}[t]{l}$P_{in}$\end{tabular}}}}%
    \put(0.77920269,0.13137576){\color[rgb]{0,0,0}\makebox(0,0)[lt]{\lineheight{1.25}\smash{\begin{tabular}[t]{l}$P_{out}$\end{tabular}}}}%
  \end{picture}%
\endgroup%

%% file: figures/powerVSA.pdf_tex
\begingroup%
  \makeatletter%
  \providecommand\color[2][]{%
    \errmessage{(Inkscape) Color is used for the text in Inkscape, but the package 'color.sty' is not loaded}%
    \renewcommand\color[2][]{}%
  }%
  \providecommand\transparent[1]{%
    \errmessage{(Inkscape) Transparency is used (non-zero) for the text in Inkscape, but the package 'transparent.sty' is not loaded}%
    \renewcommand\transparent[1]{}%
  }%
  \providecommand\rotatebox[2]{#2}%
  \newcommand*\fsize{\dimexpr\f@size pt\relax}%
  \newcommand*\lineheight[1]{\fontsize{\fsize}{#1\fsize}\selectfont}%
  \ifx\svgwidth\undefined%
    \setlength{\unitlength}{184.50117216bp}%
    \ifx\svgscale\undefined%
      \relax%
    \else%
      \setlength{\unitlength}{\unitlength * \real{\svgscale}}%
    \fi%
  \else%
    \setlength{\unitlength}{\svgwidth}%
  \fi%
  \global\let\svgwidth\undefined%
  \global\let\svgscale\undefined%
  \makeatother%
  \begin{picture}(1,0.36992274)%
    \lineheight{1}%
    \setlength\tabcolsep{0pt}%
    \put(0,0){\includegraphics[width=\unitlength,page=1]{./figures/powerVSA.pdf}}%
    \put(0.05412666,0.24860448){\color[rgb]{0,0,0}\makebox(0,0)[lt]{\lineheight{1.25}\smash{\begin{tabular}[t]{l}Power source\end{tabular}}}}%
    \put(0,0){\includegraphics[width=\unitlength,page=2]{./figures/powerVSA.pdf}}%
    \put(0.46601778,0.25046757){\color[rgb]{0,0,0}\makebox(0,0)[lt]{\lineheight{1.25}\smash{\begin{tabular}[t]{l}$M_1$\end{tabular}}}}%
    \put(0.46702573,0.11020304){\color[rgb]{0,0,0}\makebox(0,0)[lt]{\lineheight{1.25}\smash{\begin{tabular}[t]{l}$M_2$\end{tabular}}}}%
    \put(0.67450174,0.18306243){\color[rgb]{0,0,0}\makebox(0,0)[lt]{\lineheight{1.25}\smash{\begin{tabular}[t]{l}$E_s$\end{tabular}}}}%
    \put(0.89265902,0.18306237){\color[rgb]{0,0,0}\makebox(0,0)[lt]{\lineheight{1.25}\smash{\begin{tabular}[t]{l}$E_l$\end{tabular}}}}%
    \put(0.59606322,0.27551016){\color[rgb]{0,0,0}\makebox(0,0)[lt]{\lineheight{1.25}\smash{\begin{tabular}[t]{l}$P_{in_1}$\end{tabular}}}}%
    \put(0.59051957,0.08501082){\color[rgb]{0,0,0}\makebox(0,0)[lt]{\lineheight{1.25}\smash{\begin{tabular}[t]{l}$P_{in_2}$\end{tabular}}}}%
    \put(0.77920269,0.13137576){\color[rgb]{0,0,0}\makebox(0,0)[lt]{\lineheight{1.25}\smash{\begin{tabular}[t]{l}$P_{out}$\end{tabular}}}}%
  \end{picture}%
\endgroup%

%% file: friction_modeling.tex
\textcolor{black}{
\subsection{Frictional Contact Model} \label{sec:friction_model}
}
\textcolor{black}{
In the previous subsections, we have dealt with  idealized models. This serves as a basis to compare  BSA with VSA on a conceptual level. However, in real-world scenarios, the transition between the different actuator modes might be not fully instantaneous.}

\textcolor{black}{
To take this into account, we introduce a more realistic model of the switch-and-hold mechanism in this section. The mechanism is modeled as a pair of two friction-disc-clutches similar to e.g. \cite{Malzahn2018, Takeuchi21}. One clutch serves as a brake and the other one connects the spring to the link inertia. 
}

\textcolor{black}{
Each clutch can exhibit two operating states, either it is sticking or slipping. The state is determined by the relative velocity between the frames that the clutch connects. If the clutch brakes, the spring, $\dot{\psi} = 0$ must hold. Otherwise, if the spring is connected to the link, $\dot{\psi} = \dot{q}$. There is one braking and one connecting clutch per switch-and-hold mechanism each, giving a total number of four clutches. We therefore need to monitor four different relative velocities. As in\cite{HajFraj99}, we define
\begin{align*}
    g_{A} &\coloneqq \dot{\psi}_1, \  g_{B} \coloneqq \dot{\psi}_1 - \dot{q}_1, \\
    g_{C} &\coloneqq \dot{\psi}_2, \ g_{D} \coloneqq \dot{\psi}_2 - \dot{q}_2,
\end{align*}
and introduce the index sets
\begin{align*}
    \mathcal{I} &\coloneqq \{A, B, C, D\}, \\
    \mathcal{I}_s &\coloneqq \{i \in \mathcal{I} \mid g_i = 0 \}, \\
    \mathcal{I}_d &\coloneqq \{i \in \mathcal{I} \mid g_i \neq 0 \}.
\end{align*}
Here, $\mathcal{I}$ contains the indices of all four clutches, $\mathcal{I}_s$ gathers the indices of all sticking clutches and $\mathcal{I}_d$ includes the indices of all slipping clutches. Using the Jacobian
\begin{equation}
    \vect{\Gamma}_i \coloneqq \frac{\partial g_i}{\partial \vect{\xi}}, \quad i \in \mathcal{I}, \notag
\end{equation}
we can state the equations of motion for the double pendulum 
\begin{equation}
    \vect{\Pi} \ddot{\vect{\xi}} + \vect{\eta} + \vect{\tau}_k = \sum_{i \in \mathcal{I}_s} \vect{\Gamma}_i^{\mathsf{T}} {\zeta}_{s,i} + \sum_{i \in \mathcal{I}_d} \vect{\Gamma}_i^{\mathsf{T}} {\zeta}_{d,i}.
\end{equation}
In case of sticking contacts, the static contact torques ${\zeta}_{s,i}$ can be computed similar to \eqref{eq:constr_torque}:
\begin{equation}
\zeta_{s,i} = (\vect{\Gamma}_i \vect{\Pi}^{-1} \vect{\Gamma}_i^{\mathsf{T}})^{-1}\vect{\Gamma}_i \vect{\Pi}^{-1} ( -\vect{\tau}_k - \vect{\eta}), \quad i \in \mathcal{I}_s.
\end{equation}
The dynamic contact torque in case of a slipping contact is calculated from Coulomb's friction law
\begin{equation}\label{eq:dynamic_friction}
     {\zeta}_{d,i} = -\mathsf{sign}(g_i) \underbrace{\mu_d R F_{n,i}}_{\eqqcolon M_i}, \quad i \in \mathcal{I}_d. 
\end{equation}
Here, $\mu_d$ is the dynamic friction coefficient, $R$ is the effective radius and $F_{n,j}$ is the normal force of the clutch and $M_j$ is the clutch torque. \\
}

\begin{figure}[b]
	\centering
	\def\svgwidth{0.2\textwidth} 
		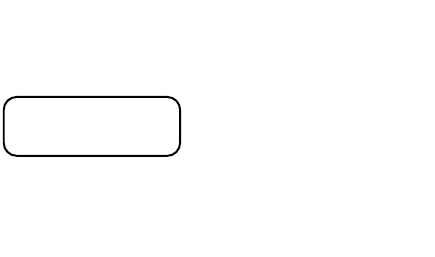
			\def\svgwidth{0.25\textwidth} 
		\hspace*{-0.5cm}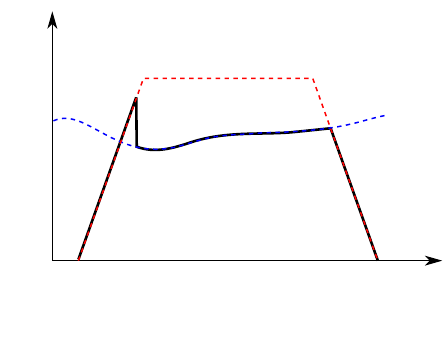
	\caption[Model overview]{Hybrid state transition and schematic plot of the friction torque. Depending on whether the contact is sticking or slipping, either $\zeta_s$ or $\zeta_d$ is applied.}
	\label{fig:friction}
\end{figure}
\textcolor{black}{
\textbf{Hybrid system model. } Again, we can formulate the dynamics as a hybrid system model. With the introduction of additional slipping modes, there are no more jumps with impulsive effects on the states. Let us first consider a single sticking contact. Since this contact is potentially slipping, we monitor the dynamic torque that the clutch can provide according to \eqref{eq:dynamic_friction}. If this torque falls below the static friction threshold, the contact begins to slip (see Fig. \ref{fig:friction}). The corresponding index is removed from $\mathcal{I}_s$ and added to $\mathcal{I}_d$. On the other hand, a slipping contact remains slipping as long as the relative velocity $g$ is unequal to zero. If there is a zero-crossing for the relative velocity and the dynamic friction torque is greater or equal to the static friction torque, the contact begins to stick again, and the respective index is removed from $\mathcal{I}_d$ and added to $\mathcal{I}_s$. 
}

\textcolor{black}{The different sticking and slipping contacts give rise to a number of different modes that are summarized in Table \ref{tab:modes_friction}. Here, we have excluded modes that contain more than three sticking contacts, since that would defy the purpose of the switch-and-hold mechanism. We define the index set $\mathcal{Q} = \{1, \dots, 9\}$, the state $\vect{x} \coloneqq [\vect{\theta}, \vect{\xi}, \dot{\vect{\xi}}]^{\mathsf{T}} \in \mathbb{R}^{10}$ and the control input $\vect{u} = \vect{u}_{\theta} \in \mathbb{R}^2$. For each mode $p \in \mathcal{Q}$, we can formulate the state space dynamics
\begin{align}
	\dot{\vect{x}} &= \vect{f}_p(\vect{x}, \vect{u}) \notag \\
	&\coloneqq \begin{bmatrix}
	\vect{u}_{\theta} \\
	\dot{\vect{\xi}} \\
	 \vect{\Pi}^{-1} ( - \vect{\eta} - \vect{\tau}_k + \underset{i \in \mathcal{I}_s^p}{\sum} \vect{\Gamma}_i^{\mathsf{T}} {\zeta}_{s,i} + \underset{i \in \mathcal{I}_d^p}{\sum} \vect{\Gamma}_j^{\mathsf{T}} {\zeta}_{d,i})
	\end{bmatrix}. \label{eq:dyn_friction}
\end{align}
Note that there are no discontinuities in the state when transitioning from one mode to another, i.e. $\vect{x}^{+} = \vect{x}^{-}$. The transitions between the modes are state-dependent. The user can open and close the clutches by providing an appropriate signal for the clutch torques $M_i$ in \eqref{eq:dynamic_friction}.  \vspace{0.5cm}
}
\begin{table}
\vspace{0.3cm}
\caption{
Index sets for the different modes of the frictional contact model.}
\centering
\renewcommand{\arraystretch}{1.4}
\color{black}{
	\begin{tabular}[c]{|l|l|}
	\hline 
	 $\mathcal{I}_s^1 = \left\{ \right\}$ & $ \mathcal{I}_d^1 = \mathcal{I}$  \\[0.06cm]
	 $\mathcal{I}_s^2 = \left\{A\right\}$ & $\mathcal{I}_d^2 = \left\{B, C, D\right\}$     \\[0.06cm]
	 $\mathcal{I}_s^3 = \left\{B\right\}$ & $\mathcal{I}_d^3 = \left\{A, C, D\right\}$     \\[0.06cm]
	 $\mathcal{I}_s^4 = \left\{C\right\}$ & $\mathcal{I}_d^4 = \left\{A, B, D\right\}$     \\[0.06cm]
	 $\mathcal{I}_s^5 = \left\{D\right\}$ & $\mathcal{I}_d^5 = \left\{A, B, C\right\}$     \\[0.06cm]
	 $\mathcal{I}_s^6 = \left\{A, C\right\}$ & $\mathcal{I}_d^6 = \left\{B, D\right\}$ \\[0.06cm]
	 $\mathcal{I}_s^7 = \left\{A, D\right\}$ & $\mathcal{I}_d^7 = \left\{B, C\right\}$ \\[0.06cm]
	 $\mathcal{I}_s^8 = \left\{B, C\right\}$ & $\mathcal{I}_d^8 = \left\{A, D\right\}$     \\[0.06cm]
     $\mathcal{I}_s^9 = \left\{B, D\right\}$ & $\mathcal{I}_d^9 = \left\{A, C\right\}$ \\[0.06cm]
 \hline
\end{tabular}
}
\label{tab:modes_friction}
\vspace{-0.50cm}
\end{table}

%% file: figures/switch.pdf_tex
\begingroup%
  \makeatletter%
  \providecommand\color[2][]{%
    \errmessage{(Inkscape) Color is used for the text in Inkscape, but the package 'color.sty' is not loaded}%
    \renewcommand\color[2][]{}%
  }%
  \providecommand\transparent[1]{%
    \errmessage{(Inkscape) Transparency is used (non-zero) for the text in Inkscape, but the package 'transparent.sty' is not loaded}%
    \renewcommand\transparent[1]{}%
  }%
  \providecommand\rotatebox[2]{#2}%
  \newcommand*\fsize{\dimexpr\f@size pt\relax}%
  \newcommand*\lineheight[1]{\fontsize{\fsize}{#1\fsize}\selectfont}%
  \ifx\svgwidth\undefined%
    \setlength{\unitlength}{125.0991557bp}%
    \ifx\svgscale\undefined%
      \relax%
    \else%
      \setlength{\unitlength}{\unitlength * \real{\svgscale}}%
    \fi%
  \else%
    \setlength{\unitlength}{\svgwidth}%
  \fi%
  \global\let\svgwidth\undefined%
  \global\let\svgscale\undefined%
  \makeatother%
  \begin{picture}(1,0.6)%
    \lineheight{1}%
    \setlength\tabcolsep{0pt}%
    \put(0,0.2){\includegraphics[width=\unitlength,page=1]{./figures/switch.pdf}}%
    \put(0.2881315,0.27624518){\color[rgb]{0,0,0}\makebox(0,0)[lt]{\lineheight{1.25}\smash{\begin{tabular}[t]{l}$\vert \zeta_d \vert \geq \vert \zeta_s \vert$\end{tabular}}}}%
    \put(0.30150202,0.66185797){\color[rgb]{0,0,0}\makebox(0,0)[lt]{\lineheight{1.25}\smash{\begin{tabular}[t]{l}$\vert \zeta_d \vert < \vert \zeta_s \vert$\end{tabular}}}}%
    \put(0.40036707,0.17025973){\color[rgb]{0,0,0}\makebox(0,0)[lt]{\lineheight{1.25}\smash{\begin{tabular}[t]{l}$g = 0$\end{tabular}}}}%
    \put(0,0.2){\includegraphics[width=\unitlength,page=2]{./figures/switch.pdf}}%
    \put(0.0451246,0.46335903){\color[rgb]{0,0,0}\makebox(0,0)[lt]{\lineheight{1.25}\smash{\begin{tabular}[t]{l}Sticking\end{tabular}}}}%
    \put(0.6,0.46258571){\color[rgb]{0,0,0}\makebox(0,0)[lt]{\lineheight{1.25}\smash{\begin{tabular}[t]{l}Slipping\end{tabular}}}}%
    \put(0,0.2){\includegraphics[width=\unitlength,page=3]{./figures/switch.pdf}}%
  \end{picture}%
\endgroup%

%% file: figures/friction_forces.pdf_tex
\begingroup%
  \makeatletter%
  \providecommand\color[2][]{%
    \errmessage{(Inkscape) Color is used for the text in Inkscape, but the package 'color.sty' is not loaded}%
    \renewcommand\color[2][]{}%
  }%
  \providecommand\transparent[1]{%
    \errmessage{(Inkscape) Transparency is used (non-zero) for the text in Inkscape, but the package 'transparent.sty' is not loaded}%
    \renewcommand\transparent[1]{}%
  }%
  \providecommand\rotatebox[2]{#2}%
  \newcommand*\fsize{\dimexpr\f@size pt\relax}%
  \newcommand*\lineheight[1]{\fontsize{\fsize}{#1\fsize}\selectfont}%
  \ifx\svgwidth\undefined%
    \setlength{\unitlength}{127.6417121bp}%
    \ifx\svgscale\undefined%
      \relax%
    \else%
      \setlength{\unitlength}{\unitlength * \real{\svgscale}}%
    \fi%
  \else%
    \setlength{\unitlength}{\svgwidth}%
  \fi%
  \global\let\svgwidth\undefined%
  \global\let\svgscale\undefined%
  \makeatother%
  \begin{picture}(1,0.7649674)%
    \lineheight{1}%
    \setlength\tabcolsep{0pt}%
    \put(0,0){\includegraphics[width=\unitlength,page=1]{./figures/friction_forces.pdf}}%
    \put(0.88278799,0.6109019){\color[rgb]{0,0,0}\makebox(0,0)[lt]{\lineheight{1.25}\smash{\begin{tabular}[t]{l}$\zeta_s$\end{tabular}}}}%
    \put(0.91993425,0.11067897){\color[rgb]{0,0,0}\makebox(0,0)[lt]{\lineheight{1.25}\smash{\begin{tabular}[t]{l}$t$\end{tabular}}}}%
    \put(0.75466659,0.05289548){\color[rgb]{0,0,0}\makebox(0,0)[lt]{\lineheight{1.25}\smash{\begin{tabular}[t]{l}$t_{separate}$\end{tabular}}}}%
    \put(0,0){\includegraphics[width=\unitlength,page=2]{./figures/friction_forces.pdf}}%
    \put(0.22138589,0.05583546){\color[rgb]{0,0,0}\makebox(0,0)[lt]{\lineheight{1.25}\smash{\begin{tabular}[t]{l}$t_{connect}$\end{tabular}}}}%
    \put(0,0){\includegraphics[width=\unitlength,page=3]{./figures/friction_forces.pdf}}%
    \put(0.8826904,0.68178455){\color[rgb]{0,0,0}\makebox(0,0)[lt]{\lineheight{1.25}\smash{\begin{tabular}[t]{l}$\zeta_d$\end{tabular}}}}%
    \put(0.03838447,0.69022391){\color[rgb]{0,0,0}\makebox(0,0)[lt]{\lineheight{1.25}\smash{\begin{tabular}[t]{l}$\zeta$\end{tabular}}}}%
    \put(0,0){\includegraphics[width=\unitlength,page=4]{./figures/friction_forces.pdf}}%
  \end{picture}%
\endgroup%

%% file: results.tex
\section{Evaluation}
\label{scc:evaluation}
To evaluate the performance of the proposed actuator, we did simulation studies for comparing bi-stiffness with variable stiffness actuation. For solving the optimal control problems, we employ a direct collocation method using a third-degree Legendre polynomial. The formulation of the optimal control problems is described in the next section.
\subsection{Optimal Control}
As in \cite{Haddadin2012}, we aim to maximize the end-link velocity of the double pendulum. We can deal with the switching by solving a multi-stage optimization problem using direct collocation \cite{bertsekas1997nonlinear}. This is similar to gait optimization in legged locomotion, which involves switching from a flight to a stance phase \cite{mombaur2005open}. The decision variables are the continuous state trajectory $\vect{x}(t)$, the control effort time-series $\vect{u}(t)$ and the duration of each mode $T_p$, which defines our switching signal. Note, that we have to select the switching sequence in advance. In each stage, we enforce the dynamics by introducing defect constraints \eqref{eq:dyn}. At time $T_p$, we initialize the next stage by evaluating the jump map in \eqref{eq:jump}. The optimal control problem can now be stated as
\begin{gather}
\underset{\vect{x}(t), \vect{u}(t), T_p}{\mathsf{min}} \notag \text{\;\;} \mathcal{J}(\vect{x}(t), \vect{u}(t)) \\
\text{s.t. \quad} \dot{\vect{x}}(t) = \vect{f}_p(\vect{x}(t), \vect{u}(t)),\;\; t \leq T_p \notag \\
\vect{x}_p^{+}(t) = \vect{g}_p(\vect{x}^{-}(t)),\;\; t = T_p \notag\\
\vect{x}(t) \in \mathcal{X}, \; \vect{u}(t) \in \mathcal{U} \notag.
\end{gather}
Here, the sets $\mathcal{X}$ and $\mathcal{U}$ indicate the constraints on the state and control effort, respectively. In case of the VSA, there is no switching and the problem reduces to finding the states and control inputs only:
\begin{gather}
\underset{\vect{x}(t), \vect{u}(t)}{\mathsf{min}} \text{\;\;} \mathcal{J}(\vect{x}(t), \vect{u}(t)) \notag \\
\text{s.t. \quad} \dot{\vect{x}}(t) = \vect{f}(\vect{x}(t), \vect{u}(t)) \notag\\
\vect{x}(t) \in \mathcal{X},\; \vect{u}(t) \in \mathcal{U} \notag.
\end{gather}
We will conduct two experiments. First, we maximize the end-link velocity $v_{TCP}$\footnote{TCP: Tool Center Point} using
\begin{equation}
    \mathcal{J}(\vect{x}(t), \vect{u}(t)) = -v_{TCP}(\vect{x}(t)) = - \vect{J}(\vect{q}(t)) \dot{\vect{q}}(t). \label{eq:cost_vel}
\end{equation}
Then, we fix the final velocity to a desired value and minimize the control effort
\begin{equation}
    \mathcal{J}(\vect{x}(t), \vect{u}(t)) = \| \vect{u}(t) \|^2. \label{eq:cost_effort}
\end{equation}
\subsection{Software}
The double pendulum was modeled using MATLAB's Symbolic Math Toolbox. The optimal control problem was formulated using CasADi \cite{Andersson2019}, and solved using Ipopt \cite{wachter2006implementation}.

\textcolor{black}{\subsection{Simulations}}
\textcolor{black}{For evaluation, we conducted three simulation experiments. First, we maximize the end-link velocity of a double pendulum, either driven by two VSAs or BSAs. Then, we fix the end-link velocity to a desired value and solve for different final times while minimizing the control effort. Finally, we apply the optimal control solution obtained in the first simulation to the model with frictional contacts.} The values of mechanical parameters for the double pendulum, except the spring inertia, are taken from \cite{Haddadin2012} and are summarized in Table \ref{tab:params}. 

The double pendulum is always initialized at its equilibrium position at $t=0$ \si{s}. The maximum motor velocity is set to $2$ rad/s for all experiments. The allowed stiffness range for the VSA is set to $K_i \in [0, 100]$ Nm/rad. The maximum stiffness adjustment rate is set to \mbox{$\dot{K}_{i, max} = 650$ Nm/(rad s)}.\footnote{Since the VSA employs a second actuator for the stiffness adjustment, we limit the stiffness adjustment for a fair comparison. We found that for a maximum stiffness adjustment rate of \mbox{$\dot{K}_{i, max} = 650$ Nm/ (rad s)}, both actuators are able to reach a final velocity of slightly below 3 \si{m/s}, when maximizing the final velocity. This way, both systems are equivalent in terms of the input power. }

\begin{figure}[t]
	\centering
	\def\svgwidth{0.52\textwidth}
	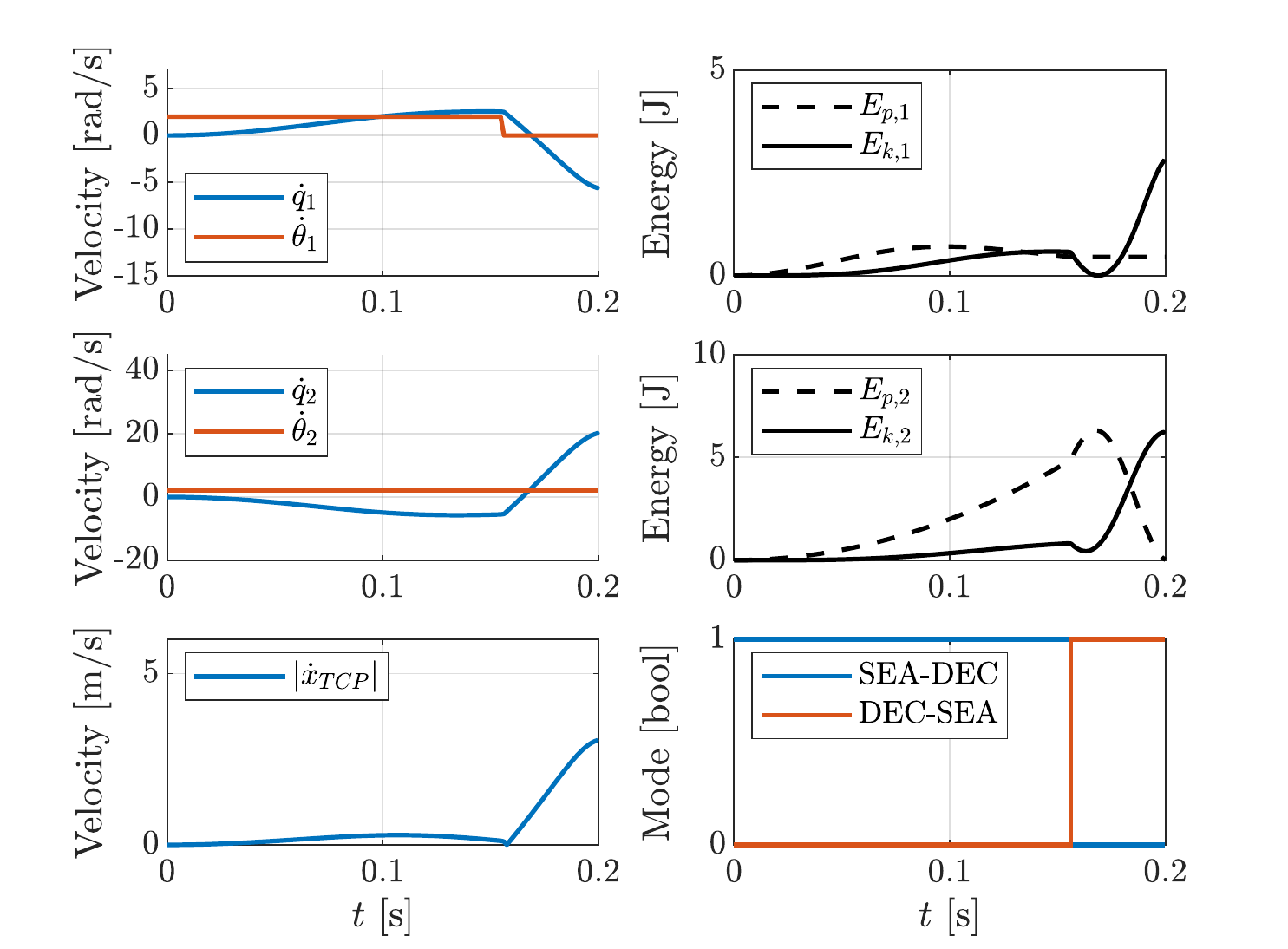 
	\caption[results]{BSA results for maximizing the end-link velocity. The kinetic and potential energy are denoted by $E_k$ and $E_p$, respectively.}
	\label{fig:DECA02}
\end{figure}

\begin{table}[ht]
\caption{Mechanical parameters}
\renewcommand{\arraystretch}{1.3}
\centering
	\begin{tabular}[c]{|c|c|c|}
	\hline	\textbf{Parameter} & \textbf{Symbol} & \textbf{Value}\\
	\hline  Mass Link 1 & $m_1$ & 5 kg\\ 
	\hline  Mass Link 2 & $m_2$ & 4.6 kg\\ 
	\hline  Moment of Inertia Link 1 & $J_{l_1}$ & 0.0453 kg m$^2$\\[0.01cm]
	\hline  Moment of Inertia Link 2 & $J_{l_2}$ & 0.0492 kg m$^2$\\[0.01cm] 
	\hline  Length Link 1 & $l_1$ & 0.34 m\\ 
	\hline  Length Link 2 & $l_2$ & 0.34 m\\ 
	\hline  Spring Inertia Joint 1 (BSA) & $J_{s_1}$ & 0.001 kg m$^2$ \\ 
	\hline  Spring Inertia Joint 2 (BSA) & $J_{s_2}$ & 0.001 kg m$^2$ \\ \hline 
\end{tabular}
\label{tab:params}
\vspace{-0.25cm}
\end{table}

\begin{figure}[t]
	\centering
	\def\svgwidth{0.52\textwidth}
	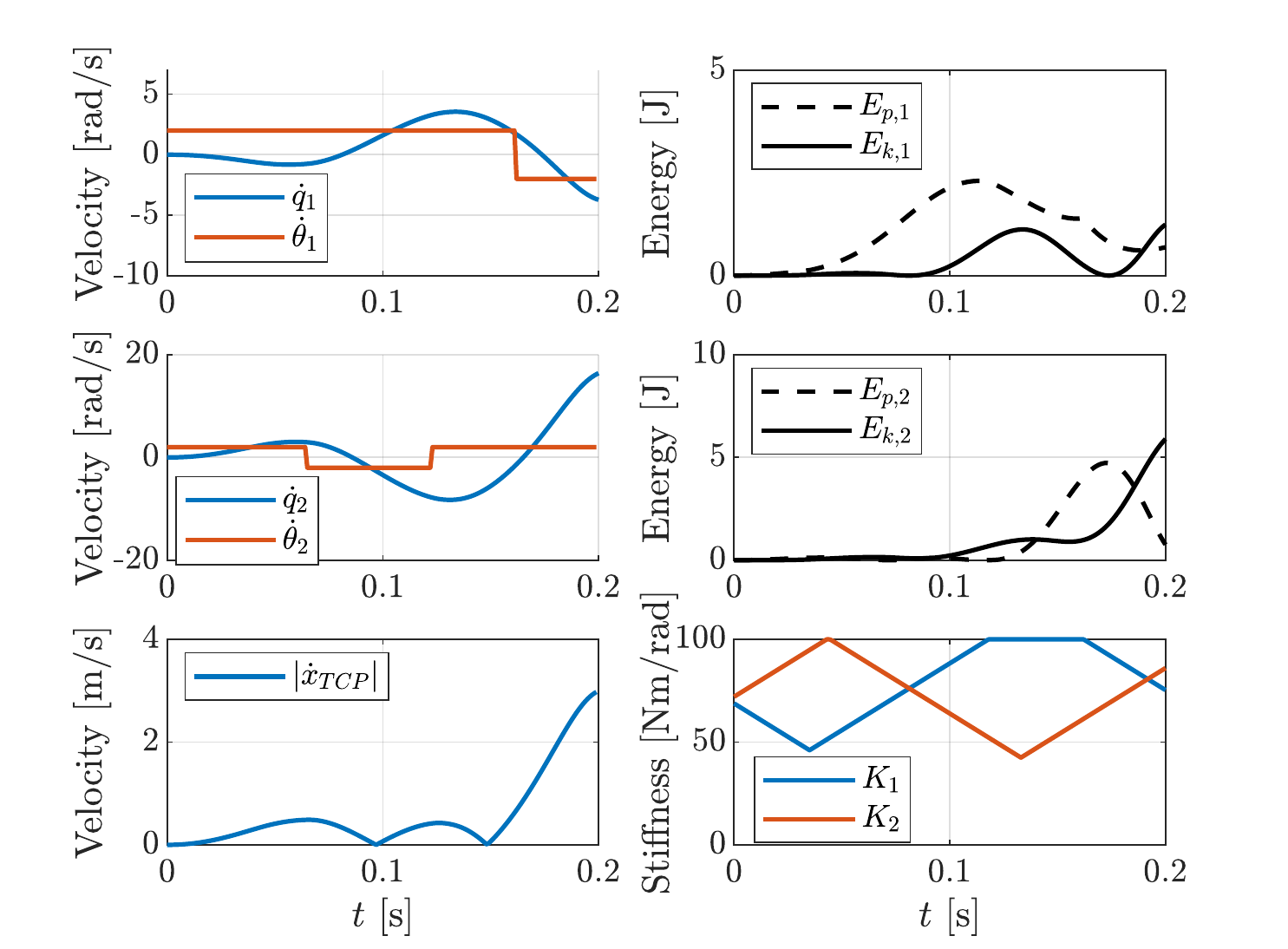 
	\caption[results]{VSA results for maximizing the end-link velocity. }
	\label{fig:VSA02_restr}
	\vspace{-0.45cm}
\end{figure}

\begin{figure}
	\centering
	\def\svgwidth{0.52\textwidth}
	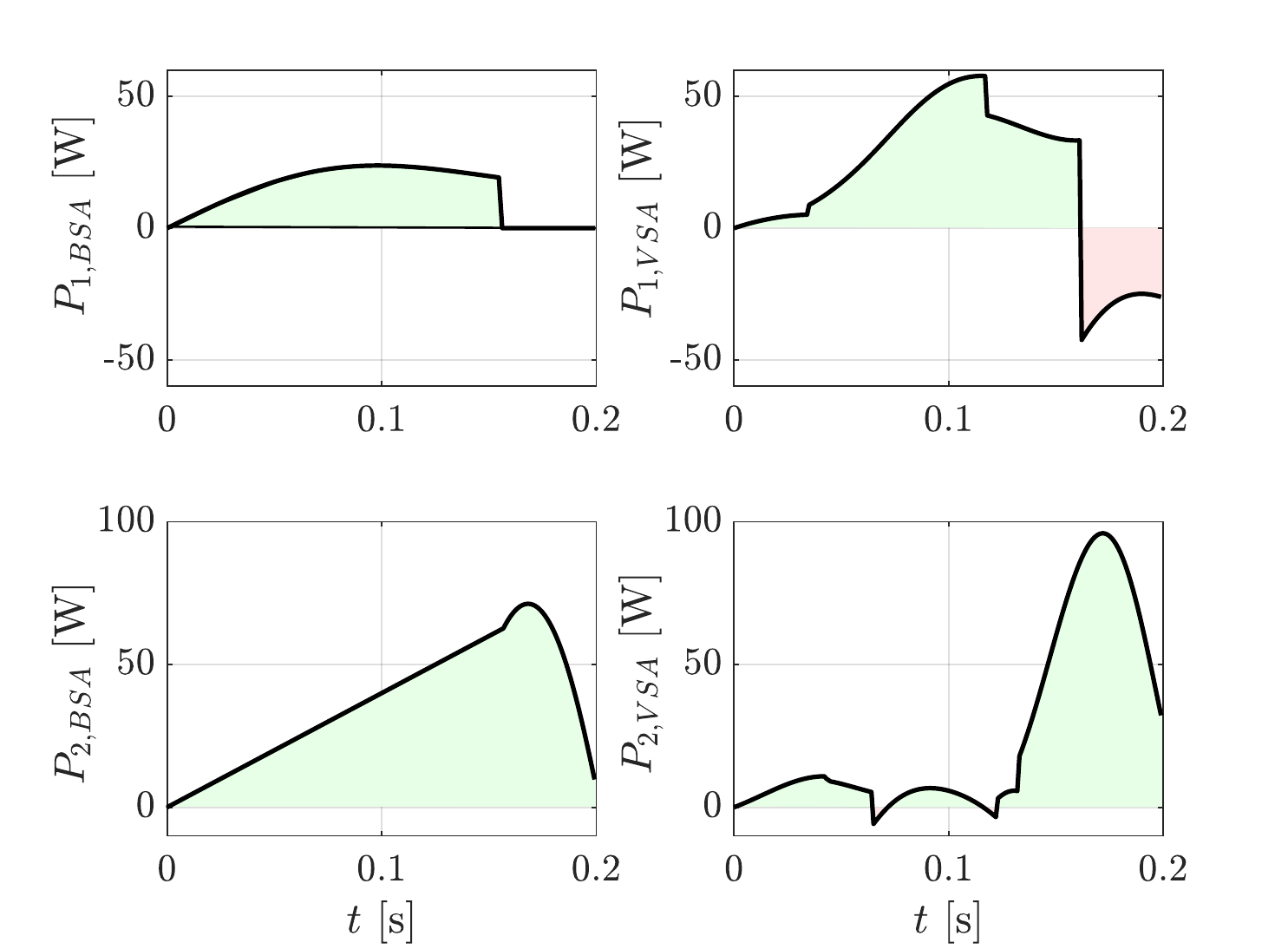 
	\vspace{-0.5cm}
		\caption[results]{Mechanical power supplied through the spring. Left: BSA. Right: VSA. The green and red areas under the curves correspond to the positive and negative work done by the spring.}\label{fig:Work}
		\vspace{-0.5cm}
\end{figure}

\begin{figure}[h]
	\centering
	\def\svgwidth{0.4\textwidth} \footnotesize
	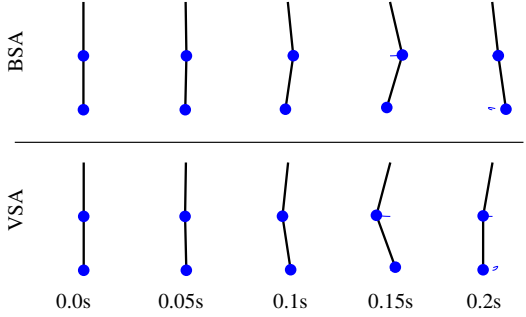
	\vspace{-0.2cm}
	\caption{Line sketch of the double pendulum's motion for BSA and VSA. Note that optimizer found a solution where the VSA throws to the left. This does not affect, however, the magnitude of the final velocity. }\label{fig:stick_figure}
	\vspace{-0.5cm}
\end{figure}

\textbf{\textcolor{black}{Simulation 1}: Maximization of end-link velocity.} 
In the first experiment, we investigate both actuators' ability to produce an explosive movement, i.e. a rapid, coordinated conversion of the stored spring energy to kinetic energy. The optimal control problem is to minimize $\mathcal{J}$ given by \eqref{eq:cost_vel} for a fixed final time of $t_f =$ 0.2 \si{s} and without state constraints. The result of the optimization for BSA and VSA are shown in Figs. \ref{fig:DECA02} and \ref{fig:VSA02_restr}, respectively. Furthermore, the power flow into the spring can be seen in Fig. \ref{fig:Work}. The motion of the double pendulum is depicted by a line sketch in Fig.~\ref{fig:stick_figure}. The VSA-driven double pendulum is able to reach a final velocity of 2.97 \si{m/s}, while the BSA-driven system is able to reach 2.99 \si{m/s}.

\textit{Observation 1: Energy transfer timing.}
For the switching sequence, we initialize the actuator such that joint 1 is in SEA mode and joint 2 is fully decoupled. The second link is therefore moved passively by being inertially coupled to the first link. During motion, the spring energy $E_{p,2}$ is continuously built up in the second joint. Just before the end of the time-interval, the launch is initiated at $T_p = 0.157$ \si{s}. Joint 2 is now coupled to the link and joint 1 is decoupled. Joint 2 rapidly transfers all of its stored spring energy in a trebuchet-like manner, resulting in the high end-link velocity.

\begin{figure}[h]
\vspace{0.3cm}
	\centering
	\def\svgwidth{0.5\textwidth}
	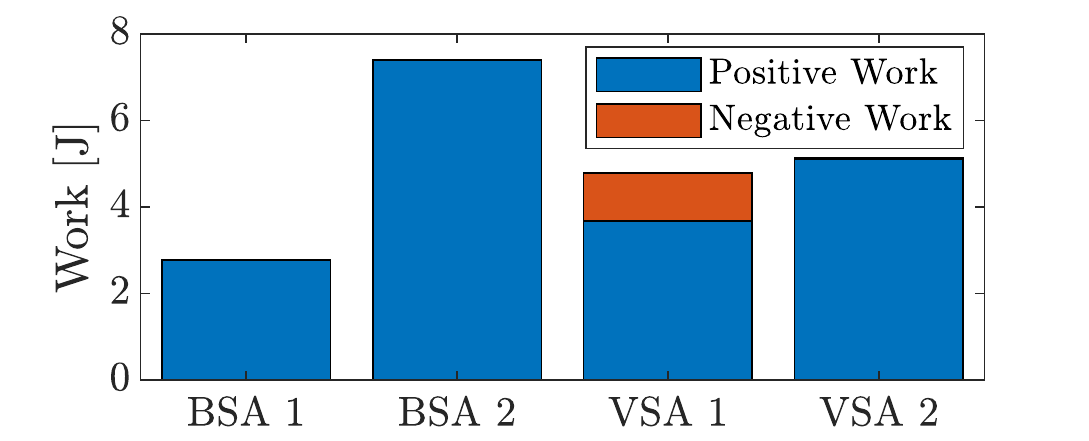 
	\caption[results]{Work done by each actuator. }\label{fig:work}
\end{figure}

\textit{Observation 2: Avoiding negative work.} The power input to the spring for each joint according to \eqref{eq:power} is shown in Fig.~\ref{fig:Work}. The amount of positive and negative work done by each actuator is summarized in Fig.~\ref{fig:work}. In total, the energy injected by each actuator is 10.16 \si{J} for the BSA and 9.92 \si{J} for the VSA. The decoupling of joint 1 after launch initiation is quite important since it would otherwise brake the movement. Since it does not contribute to the motion anymore, the power input from the motor is zero, as can be seen in Fig.~\ref{fig:Work}. Thus, there is no negative work done by the actuator.

\textit{Observation 3: VSA swing-up motion.} From Fig.~\ref{fig:VSA02_restr} and ~\ref{fig:stick_figure}, one can see that the motor velocity in the first joint closely resembles the BSA case. Moreover, the signal of the second motor switches once between the minimum and maximum motor velocity, and the stiffness signal is of a triangular shape. Furthermore, the optimizer strategy seems to be a combination of a resonant excitation plus some additional energy input by the stiffness adjustment. The resonance behavior can also be seen in the power flow plot for the second VSA joint in Fig. \ref{fig:Work}. Finally, after the first joint has transmitted its energy, the spring is actually detrimental since it brakes the motion, which can be seen in the power flow $P_{1,VSA}$ for joint 1.

\begin{figure}
	\centering
	\def\svgwidth{0.45\textwidth}
	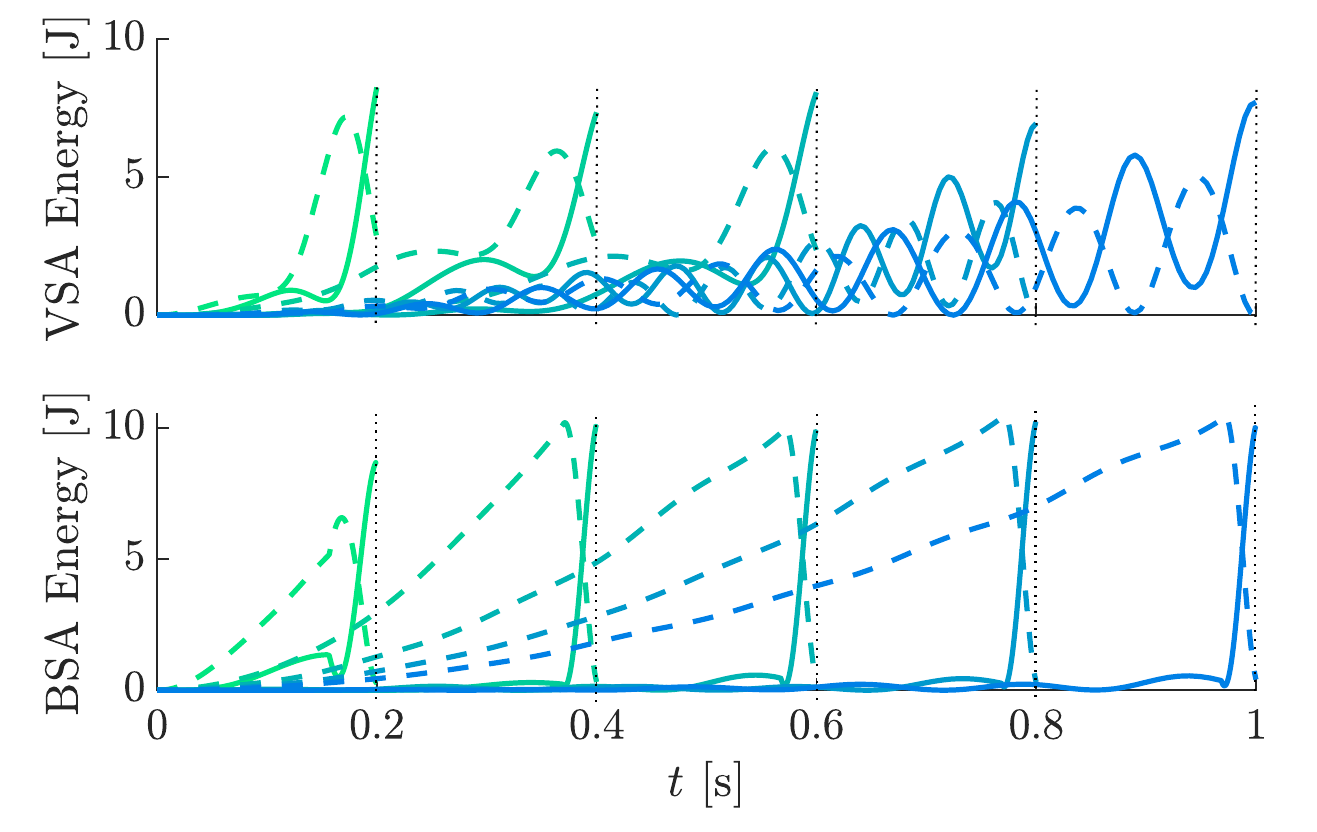
	\vspace{-0.5cm}
	\caption[results]{Total potential (dashed) and kinetic energy (solid) for different final times $t_f = 0.2 \dots 1$s (indicated by color gradient from green to blue). }\label{fig:velocities}
\end{figure}

\textbf{\textcolor{black}{Simulation 2}: Minimization of control effort for different final times.} In the second experiment, the final velocity was set to a desired value of $3$ \si{m/s}. The objective function was changed to minimize the control effort, i.e. the motor velocity and additionally, in the VSA case, the stiffness adjustment rate. We repeated the experiment for different time horizons $t_f$ ranging from 0.2 \si{s} to 1.0 \si{s}. Fig.~\ref{fig:velocities} shows the time-series of potential and kinetic energy for both systems for the respective final times.

\textit{Observation 1: Consistent BSA launch sequences.}
The BSA consistently executes a launch sequence for all final times. As can be seen in the energy plot in 
Fig.~\ref{fig:velocities}, potential energy is always continuously build-up until the end of the time-interval and then rapidly released.

\textit{Observation 2: Increasing oscillatory behavior in VSA swing up.}
In contrast, the VSA shows more and more of a resonant excitation strategy as the final time increases.  The transfer of potential energy is not timed as in the BSA case. Rather, we can observe large oscillations from potential to kinetic energy.

\textcolor{black}{
\textbf{Simulation 3: Maximization of end-link velocity using the frictional contact model. }
To further substantiate our previously obtained results, we apply the optimal control from Simulation 1 to the system with a frictional contact model according to Section \ref{sec:friction_model}. For computing the dynamic friction torque, we utilize ramp functions that increase the torque from zero to its maximum value, $M_{j, max} = 30$ Nm or vice versa. The times $t_{separate}$ and $t_{connect}$ were set to $0.02$ \si{s}. These values lie in the range of what typical electromagnetic clutches and brakes are able to achieve \cite{Mayr}. The simulation results are depicted in Fig.~\ref{fig:Fric}. To illustrate the effects of the transient dynamics better, we have added the spring velocity in the plot. In the end, the system reaches a final velocity of $2.7$ \si{m/s}. The hybrid transition graph is shown in Fig.~\ref{fig:hybrid_automaton}. \\
\textit{Observation 1: Transient slipping does not affect the energy transfer timing.}
With the frictional contact model, there is a considerable mechanical delay.  We have accounted for this fact by engaging and disengaging the respective clutches earlier than in the idealized case at $T_p=0.147$ \si{s}, i.e. $10$ \si{ms} prior. With this correction, the overall behaviour of the clutch-and-hold mechanism is not altered. It is still possible to achieve  the same clear launch sequence as in the ideal case. However, precise knowledge on the switching times of the clutches is required. \\
\begin{figure}[t]
	\centering
	\def\svgwidth{0.52\textwidth}
	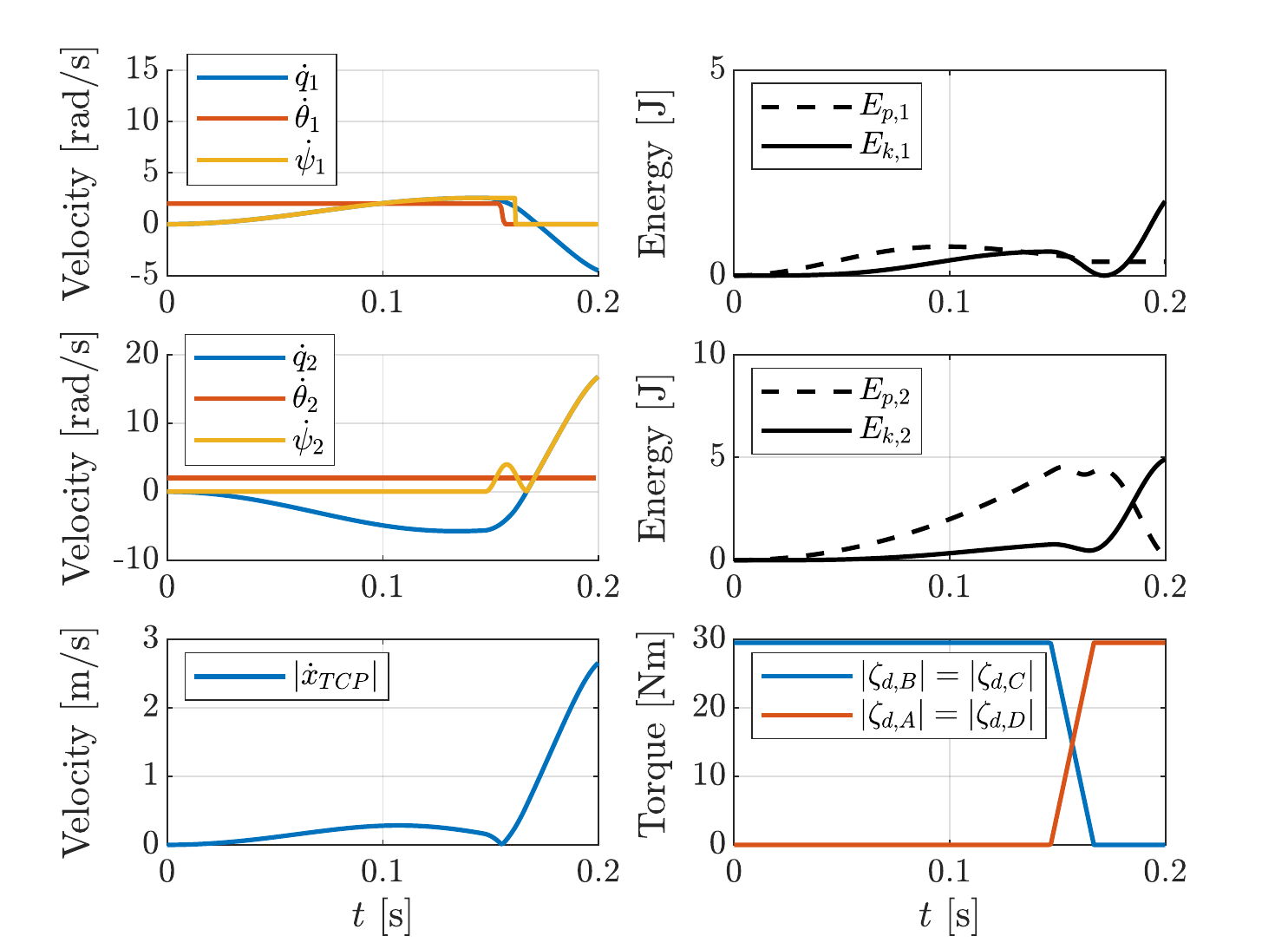 
	\caption[results]{Simulation results for the BSA system assuming a frictional contact model. }
	\label{fig:Fric}
\end{figure}
\begin{figure}[t]
	\centering
	\def\svgwidth{0.45\textwidth} \footnotesize
	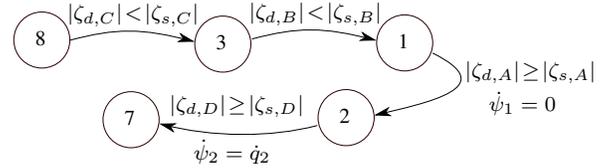 
	\caption[Model overview]{Hybrid automaton for the BSA system with a frictional contact model. The definitions of the sliding and sticking index sets are described in Table~\ref{tab:modes_friction} .}
	\label{fig:hybrid_automaton}
	\vspace{-0.5cm}
\end{figure}
\textit{Observation 2: Loss of energy due to friction.} When viewing the potential and kinetic energy plots in Fig.~\ref{fig:Fric}, it is evident that the potential energy during the slipping phase cannot be further increased. In contrast to the ideal BSA, the potential energy stored in the second spring $E_{p,2}$ remains below $5$ \si{J}. This can be explained by the fact that the clutch actually dissipates some of the kinetic energy by friction. Despite the frictional losses, we reach 90 \% of the end-link velocity in comparison to the idealized case.}
\vspace{0.25cm}

%% file: plots/DCEA.pdf_tex
\begingroup%
  \makeatletter%
  \providecommand\color[2][]{%
    \errmessage{(Inkscape) Color is used for the text in Inkscape, but the package 'color.sty' is not loaded}%
    \renewcommand\color[2][]{}%
  }%
  \providecommand\transparent[1]{%
    \errmessage{(Inkscape) Transparency is used (non-zero) for the text in Inkscape, but the package 'transparent.sty' is not loaded}%
    \renewcommand\transparent[1]{}%
  }%
  \providecommand\rotatebox[2]{#2}%
  \newcommand*\fsize{\dimexpr\f@size pt\relax}%
  \newcommand*\lineheight[1]{\fontsize{\fsize}{#1\fsize}\selectfont}%
  \ifx\svgwidth\undefined%
    \setlength{\unitlength}{427.5bp}%
    \ifx\svgscale\undefined%
      \relax%
    \else%
      \setlength{\unitlength}{\unitlength * \real{\svgscale}}%
    \fi%
  \else%
    \setlength{\unitlength}{\svgwidth}%
  \fi%
  \global\let\svgwidth\undefined%
  \global\let\svgscale\undefined%
  \makeatother%
  \begin{picture}(1,0.73684211)%
    \lineheight{1}%
    \setlength\tabcolsep{0pt}%
    \put(0,0){\includegraphics[width=\unitlength,page=1]{./plots/DCEA.pdf}}%
  \end{picture}%
\endgroup%

%% file: plots/VSA_dK_max.pdf_tex
\begingroup%
  \makeatletter%
  \providecommand\color[2][]{%
    \errmessage{(Inkscape) Color is used for the text in Inkscape, but the package 'color.sty' is not loaded}%
    \renewcommand\color[2][]{}%
  }%
  \providecommand\transparent[1]{%
    \errmessage{(Inkscape) Transparency is used (non-zero) for the text in Inkscape, but the package 'transparent.sty' is not loaded}%
    \renewcommand\transparent[1]{}%
  }%
  \providecommand\rotatebox[2]{#2}%
  \newcommand*\fsize{\dimexpr\f@size pt\relax}%
  \newcommand*\lineheight[1]{\fontsize{\fsize}{#1\fsize}\selectfont}%
  \ifx\svgwidth\undefined%
    \setlength{\unitlength}{427.5bp}%
    \ifx\svgscale\undefined%
      \relax%
    \else%
      \setlength{\unitlength}{\unitlength * \real{\svgscale}}%
    \fi%
  \else%
    \setlength{\unitlength}{\svgwidth}%
  \fi%
  \global\let\svgwidth\undefined%
  \global\let\svgscale\undefined%
  \makeatother%
  \begin{picture}(1,0.73684211)%
    \lineheight{1}%
    \setlength\tabcolsep{0pt}%
    \put(0,0){\includegraphics[width=\unitlength,page=1]{./plots/VSA_dK_max.pdf}}%
  \end{picture}%
\endgroup%

%% file: plots/Power.pdf_tex
\begingroup%
  \makeatletter%
  \providecommand\color[2][]{%
    \errmessage{(Inkscape) Color is used for the text in Inkscape, but the package 'color.sty' is not loaded}%
    \renewcommand\color[2][]{}%
  }%
  \providecommand\transparent[1]{%
    \errmessage{(Inkscape) Transparency is used (non-zero) for the text in Inkscape, but the package 'transparent.sty' is not loaded}%
    \renewcommand\transparent[1]{}%
  }%
  \providecommand\rotatebox[2]{#2}%
  \newcommand*\fsize{\dimexpr\f@size pt\relax}%
  \newcommand*\lineheight[1]{\fontsize{\fsize}{#1\fsize}\selectfont}%
  \ifx\svgwidth\undefined%
    \setlength{\unitlength}{427.5bp}%
    \ifx\svgscale\undefined%
      \relax%
    \else%
      \setlength{\unitlength}{\unitlength * \real{\svgscale}}%
    \fi%
  \else%
    \setlength{\unitlength}{\svgwidth}%
  \fi%
  \global\let\svgwidth\undefined%
  \global\let\svgscale\undefined%
  \makeatother%
  \begin{picture}(1,0.73684211)%
    \lineheight{1}%
    \setlength\tabcolsep{0pt}%
    \put(0,0){\includegraphics[width=\unitlength,page=1]{./plots/Power.pdf}}%
  \end{picture}%
\endgroup%

%% file: plots/positiveVSnegative.pdf_tex
\begingroup%
  \makeatletter%
  \providecommand\color[2][]{%
    \errmessage{(Inkscape) Color is used for the text in Inkscape, but the package 'color.sty' is not loaded}%
    \renewcommand\color[2][]{}%
  }%
  \providecommand\transparent[1]{%
    \errmessage{(Inkscape) Transparency is used (non-zero) for the text in Inkscape, but the package 'transparent.sty' is not loaded}%
    \renewcommand\transparent[1]{}%
  }%
  \providecommand\rotatebox[2]{#2}%
  \newcommand*\fsize{\dimexpr\f@size pt\relax}%
  \newcommand*\lineheight[1]{\fontsize{\fsize}{#1\fsize}\selectfont}%
  \ifx\svgwidth\undefined%
    \setlength{\unitlength}{313.5bp}%
    \ifx\svgscale\undefined%
      \relax%
    \else%
      \setlength{\unitlength}{\unitlength * \real{\svgscale}}%
    \fi%
  \else%
    \setlength{\unitlength}{\svgwidth}%
  \fi%
  \global\let\svgwidth\undefined%
  \global\let\svgscale\undefined%
  \makeatother%
  \begin{picture}(1,0.40430622)%
    \lineheight{1}%
    \setlength\tabcolsep{0pt}%
    \put(0,0){\includegraphics[width=\unitlength,page=1]{./plots/positiveVSnegative.pdf}}%
  \end{picture}%
\endgroup%

%% file: plots/animation.pdf_tex
\begingroup%
  \makeatletter%
  \providecommand\color[2][]{%
    \errmessage{(Inkscape) Color is used for the text in Inkscape, but the package 'color.sty' is not loaded}%
    \renewcommand\color[2][]{}%
  }%
  \providecommand\transparent[1]{%
    \errmessage{(Inkscape) Transparency is used (non-zero) for the text in Inkscape, but the package 'transparent.sty' is not loaded}%
    \renewcommand\transparent[1]{}%
  }%
  \providecommand\rotatebox[2]{#2}%
  \newcommand*\fsize{\dimexpr\f@size pt\relax}%
  \newcommand*\lineheight[1]{\fontsize{\fsize}{#1\fsize}\selectfont}%
  \ifx\svgwidth\undefined%
    \setlength{\unitlength}{354bp}%
    \ifx\svgscale\undefined%
      \relax%
    \else%
      \setlength{\unitlength}{\unitlength * \real{\svgscale}}%
    \fi%
  \else%
    \setlength{\unitlength}{\svgwidth}%
  \fi%
  \global\let\svgwidth\undefined%
  \global\let\svgscale\undefined%
  \makeatother%
  \begin{picture}(1,0.61864407)%
    \lineheight{1}%
    \setlength\tabcolsep{0pt}%
    \put(0,0){\includegraphics[width=\unitlength,page=1]{./plots/animation.pdf}}%
    \put(0.12060309,0.03217468){\color[rgb]{0,0,0}\makebox(0,0)[lt]{\lineheight{1.25}\smash{\begin{tabular}[t]{l}0.0s\end{tabular}}}}%
    \put(0.30768265,0.03217468){\color[rgb]{0,0,0}\makebox(0,0)[lt]{\lineheight{1.25}\smash{\begin{tabular}[t]{l}0.05s\end{tabular}}}}%
    \put(0.51802315,0.03217468){\color[rgb]{0,0,0}\makebox(0,0)[lt]{\lineheight{1.25}\smash{\begin{tabular}[t]{l}0.1s\end{tabular}}}}%
    \put(0.69239077,0.03217468){\color[rgb]{0,0,0}\makebox(0,0)[lt]{\lineheight{1.25}\smash{\begin{tabular}[t]{l}0.15s\end{tabular}}}}%
    \put(0.87307028,0.03217468){\color[rgb]{0,0,0}\makebox(0,0)[lt]{\lineheight{1.25}\smash{\begin{tabular}[t]{l}0.2s\end{tabular}}}}%
    \put(0,0){\includegraphics[width=\unitlength,page=2]{./plots/animation.pdf}}%
    \put(0.0586277,0.46626704){\color[rgb]{0,0,0}\rotatebox{90}{\makebox(0,0)[lt]{\lineheight{1.25}\smash{\begin{tabular}[t]{l}BSA\end{tabular}}}}}%
    \put(0.0586277,0.1745364){\color[rgb]{0,0,0}\rotatebox{90}{\makebox(0,0)[lt]{\lineheight{1.25}\smash{\begin{tabular}[t]{l}VSA\end{tabular}}}}}%
    \put(0,0){\includegraphics[width=\unitlength,page=3]{./plots/animation.pdf}}%
  \end{picture}%
\endgroup%

%% file: plots/energies.pdf_tex
\begingroup%
  \makeatletter%
  \providecommand\color[2][]{%
    \errmessage{(Inkscape) Color is used for the text in Inkscape, but the package 'color.sty' is not loaded}%
    \renewcommand\color[2][]{}%
  }%
  \providecommand\transparent[1]{%
    \errmessage{(Inkscape) Transparency is used (non-zero) for the text in Inkscape, but the package 'transparent.sty' is not loaded}%
    \renewcommand\transparent[1]{}%
  }%
  \providecommand\rotatebox[2]{#2}%
  \newcommand*\fsize{\dimexpr\f@size pt\relax}%
  \newcommand*\lineheight[1]{\fontsize{\fsize}{#1\fsize}\selectfont}%
  \ifx\svgwidth\undefined%
    \setlength{\unitlength}{379.03482056bp}%
    \ifx\svgscale\undefined%
      \relax%
    \else%
      \setlength{\unitlength}{\unitlength * \real{\svgscale}}%
    \fi%
  \else%
    \setlength{\unitlength}{\svgwidth}%
  \fi%
  \global\let\svgwidth\undefined%
  \global\let\svgscale\undefined%
  \makeatother%
  \begin{picture}(1,0.63210703)%
    \lineheight{1}%
    \setlength\tabcolsep{0pt}%
    \put(0,0){\includegraphics[width=\unitlength,page=1]{./plots/energies.pdf}}%
  \end{picture}%
\endgroup%

%% file: plots/results_friction.pdf_tex
\begingroup%
  \makeatletter%
  \providecommand\color[2][]{%
    \errmessage{(Inkscape) Color is used for the text in Inkscape, but the package 'color.sty' is not loaded}%
    \renewcommand\color[2][]{}%
  }%
  \providecommand\transparent[1]{%
    \errmessage{(Inkscape) Transparency is used (non-zero) for the text in Inkscape, but the package 'transparent.sty' is not loaded}%
    \renewcommand\transparent[1]{}%
  }%
  \providecommand\rotatebox[2]{#2}%
  \newcommand*\fsize{\dimexpr\f@size pt\relax}%
  \newcommand*\lineheight[1]{\fontsize{\fsize}{#1\fsize}\selectfont}%
  \ifx\svgwidth\undefined%
    \setlength{\unitlength}{427.5bp}%
    \ifx\svgscale\undefined%
      \relax%
    \else%
      \setlength{\unitlength}{\unitlength * \real{\svgscale}}%
    \fi%
  \else%
    \setlength{\unitlength}{\svgwidth}%
  \fi%
  \global\let\svgwidth\undefined%
  \global\let\svgscale\undefined%
  \makeatother%
  \begin{picture}(1,0.73684211)%
    \lineheight{1}%
    \setlength\tabcolsep{0pt}%
    \put(0,0){\includegraphics[width=\unitlength,page=1]{./plots/results_friction.pdf}}%
  \end{picture}%
\endgroup%

%% file: figures/friction_transition.pdf_tex
\begingroup%
  \makeatletter%
  \providecommand\color[2][]{%
    \errmessage{(Inkscape) Color is used for the text in Inkscape, but the package 'color.sty' is not loaded}%
    \renewcommand\color[2][]{}%
  }%
  \providecommand\transparent[1]{%
    \errmessage{(Inkscape) Transparency is used (non-zero) for the text in Inkscape, but the package 'transparent.sty' is not loaded}%
    \renewcommand\transparent[1]{}%
  }%
  \providecommand\rotatebox[2]{#2}%
  \newcommand*\fsize{\dimexpr\f@size pt\relax}%
  \newcommand*\lineheight[1]{\fontsize{\fsize}{#1\fsize}\selectfont}%
  \ifx\svgwidth\undefined%
    \setlength{\unitlength}{474.40984771bp}%
    \ifx\svgscale\undefined%
      \relax%
    \else%
      \setlength{\unitlength}{\unitlength * \real{\svgscale}}%
    \fi%
  \else%
    \setlength{\unitlength}{\svgwidth}%
  \fi%
  \global\let\svgwidth\undefined%
  \global\let\svgscale\undefined%
  \makeatother%
  \begin{picture}(1,0.2653976)%
    \lineheight{1}%
    \setlength\tabcolsep{0pt}%
    \put(0,0){\includegraphics[width=\unitlength,page=1]{./figures/friction_transition.pdf}}%
    \put(0.11429057,0.1880713){\color[rgb]{0,0,0}\makebox(0,0)[lt]{\lineheight{1.25}\smash{\begin{tabular}[t]{l}8\end{tabular}}}}%
    \put(0.85359338,0.08102445){\color[rgb]{0,0,0}\makebox(0,0)[lt]{\lineheight{1.25}\smash{\begin{tabular}[t]{l}$\dot{\psi}_1 = 0$\end{tabular}}}}%
    \put(0.3726495,0.00214343){\color[rgb]{0,0,0}\makebox(0,0)[lt]{\lineheight{1.25}\smash{\begin{tabular}[t]{l}$\dot{\psi}_2 = \dot{q}_2$\end{tabular}}}}%
    \put(0.16659479,0.22965107){\color[rgb]{0,0,0}\makebox(0,0)[lt]{\lineheight{1.25}\smash{\begin{tabular}[t]{l}$|\zeta_{d, C}|\!<\!|\zeta_{s, C}|$\end{tabular}}}}%
    \put(0,0){\includegraphics[width=\unitlength,page=2]{./figures/friction_transition.pdf}}%
    \put(0.40910388,0.18140563){\color[rgb]{0,0,0}\makebox(0,0)[lt]{\lineheight{1.25}\smash{\begin{tabular}[t]{l}3\end{tabular}}}}%
    \put(0.46285657,0.23160605){\color[rgb]{0,0,0}\makebox(0,0)[lt]{\lineheight{1.25}\smash{\begin{tabular}[t]{l}$|\zeta_{d, B}|\!<\!|\zeta_{s, B}|$\end{tabular}}}}%
    \put(0,0){\includegraphics[width=\unitlength,page=3]{./figures/friction_transition.pdf}}%
    \put(0.70536442,0.18336058){\color[rgb]{0,0,0}\makebox(0,0)[lt]{\lineheight{1.25}\smash{\begin{tabular}[t]{l}1\end{tabular}}}}%
    \put(0.81490194,0.13689864){\color[rgb]{0,0,0}\makebox(0,0)[lt]{\lineheight{1.25}\smash{\begin{tabular}[t]{l}$|\zeta_{d, A}|\!\geq\!|\zeta_{s, A}|$\end{tabular}}}}%
    \put(0,0){\includegraphics[width=\unitlength,page=4]{./figures/friction_transition.pdf}}%
    \put(0.60995157,0.0615616){\color[rgb]{0,0,0}\makebox(0,0)[lt]{\lineheight{1.25}\smash{\begin{tabular}[t]{l}2\end{tabular}}}}%
    \put(0,0){\includegraphics[width=\unitlength,page=5]{./figures/friction_transition.pdf}}%
    \put(0.25964231,0.05815641){\color[rgb]{0,0,0}\makebox(0,0)[lt]{\lineheight{1.25}\smash{\begin{tabular}[t]{l}7\end{tabular}}}}%
    \put(0.33343116,0.08197591){\color[rgb]{0,0,0}\makebox(0,0)[lt]{\lineheight{1.25}\smash{\begin{tabular}[t]{l}$|\zeta_{d, D}|\!\geq\!|\zeta_{s, D}|$\end{tabular}}}}%
    \put(0,0){\includegraphics[width=\unitlength,page=6]{./figures/friction_transition.pdf}}%
  \end{picture}%
\endgroup%

%% file: conclusion.tex
\section{Conclusion}
\label{scc:conc}
In this paper, we proposed a Bi-Stiffness Actuation concept incorporating a switch-and-hold clutch mechanism with which it is possible to fully decouple the link from the joint mechanism while keeping the elastic energy stored.
Inspired by explosive movements like throwing in biological systems, 
we defined a control sequence switching between coupled (being equivalent to a SEA) and decoupled modes, which resembles the proximo-distal sequence in biomechanics. 
By doing this, the energy transfer timing can be intuitively controlled and directly optimized for when maximizing the end-link velocity. Our numerical experiments demonstrated that a BSA can achieve similar performance compared to a power-equivalent VSA. 
Furthermore, our investigation showed that with VSAs, the optimal behavior would inevitably increase the oscillatory amplitude. In contrast, our proposed mechanism can produce a clear launch sequence for a desired end-link velocity with varying final times.
This makes the energy storage and release to be intuitively controllable, and exposes transfer timing to be an explicitly optimizable parameter.
\textcolor{black}{We were also able to show that even when considering a frictional contact model that enabled transient slipping modes during switching, we were able to achieve a similar launch sequence and energy transfer timing as in the ideal case.}
In the future, we aim to realize a first prototype of the proposed actuator concept. Another important extension is the application to a more complex robot model to examine how the results of energy transfer timing can also be applied to this case.